\documentclass{article}

\usepackage[nonatbib,final]{nips_2016}

\usepackage[utf8]{inputenc} 
\usepackage[T1]{fontenc}    
\usepackage{hyperref}       
\usepackage{url}            
\usepackage{booktabs}       
\usepackage{amsfonts}       
\usepackage{nicefrac}       
\usepackage{microtype}      

\usepackage{graphicx} 
\usepackage{widetext}
\usepackage{amsmath}
\usepackage{amssymb}
\usepackage{amsthm}
\usepackage{algorithm}
\usepackage{algorithmic}

\newcommand{\safemath}[2]{\newcommand{#1}{\ensuremath{#2}\xspace}}

\DeclareMathOperator*{\argmin}{arg\,min}
\DeclareMathOperator*{\argmax}{arg\,max}

\newcommand{\R}{\mathbb{R}}

\safemath{\Ab}{\mathbb{A}}
\safemath{\Bb}{\mathbb{B}}
\safemath{\Cb}{\mathbb{C}}
\safemath{\Db}{\mathbb{D}}
\safemath{\Eb}{\mathbb{E}}
\safemath{\Fb}{\mathbb{F}}
\safemath{\Gb}{\mathbb{G}}
\safemath{\Hb}{\mathbb{H}}
\safemath{\Ib}{\mathbb{I}}
\safemath{\Jb}{\mathbb{J}}
\safemath{\Kb}{\mathbb{K}}
\safemath{\Lb}{\mathbb{L}}
\safemath{\Mb}{\mathbb{M}}
\safemath{\Nb}{\mathbb{N}}
\safemath{\Ob}{\mathbb{O}}
\safemath{\Pb}{\mathbb{P}}
\safemath{\Qb}{\mathbb{Q}}
\safemath{\Rb}{\mathbb{R}}
\safemath{\Sb}{\mathbb{S}}
\safemath{\Tb}{\mathbb{T}}
\safemath{\Ub}{\mathbb{U}}
\safemath{\Vb}{\mathbb{V}}
\safemath{\Wb}{\mathbb{W}}
\safemath{\Xb}{\mathbb{X}}
\safemath{\Yb}{\mathbb{Y}}
\safemath{\Zb}{\mathbb{Z}}

\safemath{\Af}{\mathbf{A}}
\safemath{\Bf}{\mathbf{B}}
\safemath{\Df}{\mathbf{D}}
\safemath{\Ef}{\mathbf{E}}
\safemath{\Ff}{\mathbf{F}}
\safemath{\Gf}{\mathbf{G}}
\safemath{\Hf}{\mathbf{H}}
\safemath{\Jf}{\mathbf{J}}
\safemath{\Kf}{\mathbf{K}}
\safemath{\Lf}{\mathbf{L}}
\safemath{\Mf}{\mathbf{M}}
\safemath{\Nf}{\mathbf{N}}
\safemath{\Of}{\mathbf{O}}
\safemath{\Pf}{\mathbf{P}}
\safemath{\Qf}{\mathbf{Q}}
\safemath{\Rf}{\mathbf{R}}
\safemath{\Sf}{\mathbf{S}}
\safemath{\Tf}{\mathbf{T}}
\safemath{\Uf}{\mathbf{U}}
\safemath{\Vf}{\mathbf{V}}
\safemath{\Wf}{\mathbf{W}}
\safemath{\Xf}{\mathbf{X}}
\safemath{\Yf}{\mathbf{Y}}
\safemath{\Zf}{\mathbf{Z}}

\safemath{\af}{\mathbf{a}}
\safemath{\df}{\mathbf{d}}
\safemath{\ef}{\mathbf{e}}
\safemath{\ff}{\mathbf{f}}
\safemath{\gf}{\mathbf{g}}
\safemath{\hf}{\mathbf{h}}
\safemath{\jf}{\mathbf{j}}
\safemath{\kf}{\mathbf{k}}
\safemath{\lf}{\mathbf{l}}
\safemath{\mf}{\mathbf{m}}
\safemath{\nf}{\mathbf{n}}
\safemath{\of}{\mathbf{o}}
\safemath{\pf}{\mathbf{p}}
\safemath{\qf}{\mathbf{q}}
\safemath{\rf}{\mathbf{r}}
\safemath{\tf}{\mathbf{t}}
\safemath{\uf}{\mathbf{u}}
\safemath{\vf}{\mathbf{v}}
\safemath{\wf}{\mathbf{w}}
\safemath{\xf}{\mathbf{x}}
\safemath{\yf}{\mathbf{y}}
\safemath{\zf}{\mathbf{z}}

\safemath{\Ac}{\mathcal{A}}
\safemath{\Bc}{\mathcal{B}}
\safemath{\Cc}{\mathcal{C}}
\safemath{\Dc}{\mathcal{D}}
\safemath{\Ec}{\mathcal{E}}
\safemath{\Fc}{\mathcal{F}}
\safemath{\Gc}{\mathcal{G}}
\safemath{\Hc}{\mathcal{H}}
\safemath{\Ic}{\mathcal{I}}
\safemath{\Jc}{\mathcal{J}}
\safemath{\Kc}{\mathcal{K}}
\safemath{\Lc}{\mathcal{L}}
\safemath{\Mc}{\mathcal{M}}
\safemath{\Nc}{\mathcal{N}}
\safemath{\Oc}{\mathcal{O}}
\safemath{\Pc}{\mathcal{P}}
\safemath{\Qc}{\mathcal{Q}}
\safemath{\Rc}{\mathcal{R}}
\safemath{\Sc}{\mathcal{S}}
\safemath{\Tc}{\mathcal{T}}
\safemath{\Uc}{\mathcal{U}}
\safemath{\Vc}{\mathcal{V}}
\safemath{\Wc}{\mathcal{W}}
\safemath{\Xc}{\mathcal{X}}
\safemath{\Yc}{\mathcal{Y}}
\safemath{\Zc}{\mathcal{Z}}

\safemath{\As}{\mathsf{A}}
\safemath{\Bs}{\mathsf{B}}
\safemath{\Cs}{\mathsf{C}}
\safemath{\Ds}{\mathsf{D}}
\safemath{\Es}{\mathsf{E}}
\safemath{\Fs}{\mathsf{F}}
\safemath{\Gs}{\mathsf{G}}
\safemath{\Hs}{\mathsf{H}}
\safemath{\Is}{\mathsf{I}}
\safemath{\Js}{\mathsf{J}}
\safemath{\Ks}{\mathsf{K}}
\safemath{\Ls}{\mathsf{L}}
\safemath{\Ms}{\mathsf{M}}
\safemath{\Ns}{\mathsf{N}}
\safemath{\Os}{\mathsf{O}}
\safemath{\Ps}{\mathsf{P}}
\safemath{\Qs}{\mathsf{Q}}
\safemath{\Rs}{\mathsf{R}}
\safemath{\Ss}{\mathsf{S}}
\safemath{\Ts}{\mathsf{T}}
\safemath{\Us}{\mathsf{U}}
\safemath{\Vs}{\mathsf{V}}
\safemath{\Ws}{\mathsf{W}}
\safemath{\Xs}{\mathsf{X}}
\safemath{\Ys}{\mathsf{Y}}
\safemath{\Zs}{\mathsf{Z}}

\safemath{\as}{\mathsf{a}}
\safemath{\bs}{\mathsf{b}}
\safemath{\cs}{\mathsf{c}}
\safemath{\ds}{\mathsf{d}}
\safemath{\es}{\mathsf{e}}
\safemath{\fs}{\mathsf{f}}
\safemath{\gs}{\mathsf{g}}
\safemath{\hs}{\mathsf{h}}
\safemath{\is}{\mathsf{i}}
\safemath{\js}{\mathsf{j}}
\safemath{\ks}{\mathsf{k}}
\safemath{\ls}{\mathsf{l}}
\safemath{\ms}{\mathsf{m}}
\safemath{\ns}{\mathsf{n}}
\safemath{\os}{\mathsf{o}}
\safemath{\ps}{\mathsf{p}}
\safemath{\qs}{\mathsf{q}}
\safemath{\rs}{\mathsf{r}}
\safemath{\ts}{\mathsf{t}}
\safemath{\us}{\mathsf{u}}
\safemath{\xs}{\mathsf{x}}
\safemath{\ys}{\mathsf{y}}
\safemath{\zs}{\mathsf{z}}

\DeclareMathAlphabet{\mathbfsf}{\encodingdefault}{\sfdefault}{bx}{n}
\safemath{\Asf}{\mathbfsf{A}}
\safemath{\Bsf}{\mathbfsf{B}}
\safemath{\Csf}{\mathbfsf{C}}
\safemath{\Dsf}{\mathbfsf{D}}
\safemath{\Esf}{\mathbfsf{E}}
\safemath{\Fsf}{\mathbfsf{F}}
\safemath{\Gsf}{\mathbfsf{G}}
\safemath{\Hsf}{\mathbfsf{H}}
\safemath{\Isf}{\mathbfsf{I}}
\safemath{\Jsf}{\mathbfsf{J}}
\safemath{\Ksf}{\mathbfsf{K}}
\safemath{\Lsf}{\mathbfsf{L}}
\safemath{\Msf}{\mathbfsf{M}}
\safemath{\Nsf}{\mathbfsf{N}}
\safemath{\Osf}{\mathbfsf{O}}
\safemath{\Psf}{\mathbfsf{P}}
\safemath{\Qsf}{\mathbfsf{Q}}
\safemath{\Rsf}{\mathbfsf{R}}
\safemath{\Ssf}{\mathbfsf{S}}
\safemath{\Tsf}{\mathbfsf{T}}
\safemath{\Usf}{\mathbfsf{U}}
\safemath{\Vsf}{\mathbfsf{V}}
\safemath{\Wsf}{\mathbfsf{W}}
\safemath{\Xsf}{\mathbfsf{X}}
\safemath{\Ysf}{\mathbfsf{Y}}
\safemath{\Zsf}{\mathbfsf{Z}}

\safemath{\asf}{\mathbfsf{a}}
\safemath{\bsf}{\mathbfsf{b}}
\safemath{\csf}{\mathbfsf{c}}
\safemath{\dsf}{\mathbfsf{d}}
\safemath{\esf}{\mathbfsf{e}}
\safemath{\fsf}{\mathbfsf{f}}
\safemath{\gsf}{\mathbfsf{g}}
\safemath{\hsf}{\mathbfsf{h}}
\safemath{\isf}{\mathbfsf{i}}
\safemath{\jsf}{\mathbfsf{j}}
\safemath{\ksf}{\mathbfsf{k}}
\safemath{\lsf}{\mathbfsf{l}}
\safemath{\msf}{\mathbfsf{m}}
\safemath{\nsf}{\mathbfsf{n}}
\safemath{\osf}{\mathbfsf{o}}
\safemath{\psf}{\mathbfsf{p}}
\safemath{\qsf}{\mathbfsf{q}}
\safemath{\rsf}{\mathbfsf{r}}
\safemath{\ssf}{\mathbfsf{s}}
\safemath{\tsf}{\mathbfsf{t}}
\safemath{\usf}{\mathbfsf{u}}
\safemath{\vsf}{\mathbfsf{v}}
\safemath{\wsf}{\mathbfsf{w}}
\safemath{\xsf}{\mathbfsf{x}}
\safemath{\ysf}{\mathbfsf{y}}
\safemath{\zsf}{\mathbfsf{z}}

\newtheorem{lemma}{Lemma}

\newcommand{\etal}{\mbox{\emph{et al.\ }}}

\usepackage{times}
\title{k$^2$-means for fast and accurate large scale clustering}

\author{
  Eirikur Agustsson\\
  Computer Vision Lab\\
  D-ITET\\
  ETH Zurich\\
  \texttt{aeirikur@vision.ee.ethz.ch} \\
  \And
  Radu Timofte \\
  Computer Vision Lab\\
  D-ITET\\
  ETH Zurich\\
  \texttt{timofter@vision.ee.ethz.ch} \\
  \AND
  Luc Van Gool \\
  ESAT, KU Leuven \\
  D-ITET, ETH Zurich\\
  \texttt{vangool@vision.ee.ethz.ch} \\
}

\begin{document}

\maketitle

\begin{abstract}
We propose $k^2$-means, a new clustering method which efficiently copes with large numbers of clusters and achieves low energy solutions.
$k^2$-means builds upon the standard $k$-means (Lloyd's algorithm) and combines a new strategy to accelerate the convergence with a new low time complexity divisive initialization. The accelerated convergence is achieved through only looking at $k_n$ nearest clusters and using triangle inequality bounds in the assignment step while the divisive initialization employs an optimal 2-clustering along a direction.
The worst-case time complexity per iteration of our $k^2$-means is $O(nk_nd+k^2d)$, where $d$ is the dimension of the $n$ data points and $k$ is the number of clusters and usually $n\gg k \gg k_n$. 
Compared to $k$-means' $O(nkd)$ complexity, our $k^2$-means complexity is significantly lower, at the expense of slightly increasing the memory complexity by $O(nk_n+k^2)$.
In our extensive experiments $k^2$-means is order(s) of magnitude faster than standard methods in computing accurate clusterings on several standard datasets and settings with hundreds of clusters and high dimensional data.
Moreover, the proposed divisive initialization generally leads to clustering energies comparable to those achieved with the standard $k$-means++ initialization, while being significantly faster.
\end{abstract}

\section{Introduction}
\label{sec:intro}

The $k$-means algorithm in its standard form (Lloyd's algorithm) employs two steps to cluster $n$ data points of $d$ dimensions and $k$ initial cluster centers~\cite{lloyd1982least}. The \textit{expectation} or \textit{assignment step} assigns each point to its nearest cluster while the \textit{maximization} or \textit{update step} updates the $k$ cluster centers with the mean of the points belonging to each cluster. The $k$-means algorithm repeats the two steps until convergence, that is the assignments no longer change in an iteration $i$.

$k$-means is one of the most widely used clustering algorithms, being included in a list of top 10 data mining algorithms~\cite{Wu-KIS-2008}. Its simplicity and general applicability vouch for its broad adoption. 
Unfortunately, its $O(ndki)$ time complexity depends on the product between number of points $n$, number of dimensions $d$, number of clusters $k$, and number of iterations $i$. Thus, for large such values even a single iteration of the algorithm is very slow.

The simplest way to handle larger datasets is parallelization~\cite{Zhao-CC-2009,Xu-PDS-2014}, however this requires more computation power as well. Another way is to process the data online in batches as done by the MiniBatch algorithm of Sculley~\cite{Sculley-WWW-2010}, a variant of the Lloyd algorithm that trades off quality (i.e. the converged energy) for speed.

To improve both the speed and the quality of the clustering results, Arthur and Vassilvitskii~\cite{Arthur-DA-2007} proposed the $k$-means++ initialization method. The initialization typically results in a higher quality clustering and fewer iterations for $k$-means, than when using the default random initialization. Furthermore, the expected value of the clustering energy is within a $8(\ln k +2)$ factor of the optimal solution.
However, the time complexity of the method is $O(ndk)$, i.e. the same as a single iteration of the Lloyd algorithm - which can be too expensive in a large scale setting.
Since $k$-means++ is sequential in nature, Bahman~\etal~\cite{Bahmani-VLDBE-2012} introduced a parallel version $k$-means$||$ of $k$-means++, but did not reduce the time complexity of the method.

Another direction is to speed up the actual $k$-means iterations. Elkan~\cite{elkan2003using}, Hamerly~\cite{hamerly2010making} and Drake and Hamerly~\cite{drake2012accelerated} go in this direction and use the triangle inequality to avoid unnecessary distance computation between cluster centers and the data points. However, these methods still require a full Lloyd iteration in the beginning to then gradually reduce the computation of progressive iterations.
The recent Yinyang $k$-means method of Ding~\etal~\cite{Ding-ICML-2015} is a similar method, that also leverages bounds to avoid redundant distance calculations. While typically performing  2-3$\times$ faster than Elkan method, it also requires a full Lloyd iteration to start with.

Philbin~\etal~\cite{philbin2007object} introduce an approximate $k$-means (AKM) method based on kd-trees to speed up the assignment step, reducing the complexity of each $k$-means iteration from $O(nkd)$ to $O(nmd)$, where $m<k$. In this case $m$, the distance computations performed per each iteration, controls the trade-off between a fast and an accurate (i.e. low energy) clustering. Wang~\etal~\cite{wang2012fast} use cluster closures for further $2.5\times$ speedups.

In this paper we propose $k^2$-means, a method aiming at both fast and accurate clustering.
Following the observation that usually the clusters change gradually and affect only local neighborhoods,
in the assignment step we only consider the $k_n$ nearest neighbours of a center as the candidates for the clusters members.
Furthermore we employ the triangle inequality bounds idea as introduced by Elkan~\cite{elkan2003using}  to reduce the number of operations per each iteration.
For initializing $k^2$-means, we propose a divisive initialization method, which we experimentally prove to be more efficient than $k$-means++.

Our $k^2$-means gives a significant algorithmic speedup, i.e. reducing the complexity to $O(nk_nd)$ per iteration,  while still maintaining a high accuracy comparable to methods such as $k$-means++ for a chosen $k_n < k$.
Similar to $m$ in AKM, $k_n$ also controls a trade-off between speed and accuracy. However, our experiments show that we can use a significantly lower $k_n$ when aiming for a high accuracy.

The paper is structured as follows. In Table~\ref{tab:notations} we summarize the notations used in this paper. In Section~\ref{sec:k2means} we introduce our proposed $k^2$-means method and our divisive initialization. In Section~\ref{sec:experiments} we describe the experimental benchmark and discuss the results obtained, while in Section~\ref{sec:conclusions} we draw conclusions.

\begin{minipage}[c]{0.48\linewidth}
\begin{algorithm}[H]
\caption{$k^2$-means}\label{alg:k2means}
\begin{algorithmic}[1]
\small
\STATE \textbf{Given}: $k$, data $X$, neighbourhood size $k_n$
\STATE Initialize centers $C$
\STATE Initialize assignments $a: \{1 \cdots n \} \rightarrow \{1,\cdots, k\}$ .
\WHILE{Not converged}
\STATE Build $k_n$-NN graph of $C$: \label{buildnngraph}
\STATE $\mathcal{N}_{k_n}: C \rightarrow \{1\cdots k\}^{k_n}$
\FOR{$x\in X$}
\STATE Get current center for $x$:
\STATE $l \gets a(x)$
\STATE Assign $x$ to nearest candidate center:
\STATE $a(x) \gets  \argmin_{l'\in \mathcal{N}_{k_n}(c_l)} \|x-c_{l'}\|$\label{canddist}
\ENDFOR
\FOR{$j\in \{ 1 \cdots k\}$}
\STATE $c_j \gets \mu(X_j)$
\COMMENT{Update center}
\ENDFOR
\ENDWHILE
\STATE \textbf{return} $C,a$
\end{algorithmic}
\end{algorithm}
\vspace{-0.5cm}
\end{minipage}\hfill  
\begin{minipage}[c]{0.48\linewidth}
\begin{table}[H]\label{tab:notations}
\caption{Notations}
\begin{center}
\small
\tabcolsep=0.15cm
\resizebox{\linewidth}{!}{
\begin{tabular}{l l}
\hline
$n$ & number of data points to cluster\\
$k$ & number of clusters\\
$k_n$ & number of nearest clusters\\
$d$ & number of dimensions of the data points\\
$X$ & the data $(x_i)_{i=1}^n, x_i \in \mathbb{R}^d$\\
$C$ & cluster centers $C=(c_j)_{j=1}^k, c_j \in  \mathbb{R}^d$\\
$a$ & cluster assignments $\{1,\cdots,n\} \rightarrow \{1,\cdots,k\}$\\
$a(x_i)$ & cluster assignment of $x_i$, i.e. $a(i)$ \\
$a(X')$ & cluster assignment of some set of points $X'$ \\
$X_j$ & points assigned to cluster $j$, $(x_i\in X | a(i) = j)$ \\
$\mu(X_j)$ & the mean of $X_j$: $\frac{1}{|X_j|} \sum_{x\in X_j}x$\\
$\| x \|$ & $l_2$ norm of $x\in\R^d$\\
$\phi(X_j)$ & energy of $X_j$: $\sum_{x\in X_j} \| x-\mu(X_j)\|^2$\\
$\mathcal{N}_{k_n}(c_l)$ & $k_n$ nearest neighbours of $c_l$ in $C$ (including $c_l$)\\
\hline
\end{tabular}
}
\end{center}
\vspace{-0.5cm}
\end{table}
\end{minipage}\hfill

\section{Proposed $k^2$-means}
\label{sec:k2means}

In this section we introduce our $k^2$-means method and motivate the design decisions. The pseudocode of the method is given in Algorithm~\ref{alg:k2means}. 

Given some data $X=(x_i)_{i=1}^n, x_i \in \mathbb{R}^d$, 
the $k$-means clustering objective is to find cluster centers $C=(c_j)_{j=1}^k, c_j \in  \mathbb{R}^d$ and cluster assignments $a: \{1,\cdots,n\} \rightarrow \{1,\cdots,k\}$, such that the cluster energy
\begin{equation}
\sum_{j=1}^k \sum_{ x \in X_j } \|x-c_j\|^2
\end{equation}
is minimized, 
where $X_j := (x_i\in X | a(i) = j)$ denotes the points assigned to a cluster $j$.
For a data point $x_i$, we sometimes write $a(x_i)$ instead of $a(i)$ for the cluster assignment. Similarly, for a subset $X'$ of the data, $a(X')$ denotes the cluster assignments of the corresponding points.

Standard Lloyd obtains an approximate solution by repeating the following until convergence: 
i) In the \textit{assignment step}, each $x$ is assigned to the nearest center in $C$. ii) For the \textit{update step}, each center is recomputed as the mean of its members.

The assignment step requires $O(nk)$ distance computations, i.e. $O(nkd)$ operations, and dominates the time complexity of each iteration. The update step requires only $O(nd)$ operations for mean computations.

To speed up the assignment step, an approximate nearest neighbour method can be used, such as kd-trees~\cite{philbin2007object,muja2009fast} or locality sensitive hashing~\cite{indyk1998approximate}. However, these methods ignore the fact that the cluster centers \textit{are moving} across iterations and often this movement is \textit{slow}, affecting a \textit{small neighborhood} of points. With this observation, we obtain a very simple fast nearest neighbour scheme:

Suppose at iteration $i$, a data point $x$ was assigned to a nearby center, $l=a(x)$.
After updating the centers, we still expect $c_l$ to be close to $x$.
Therefore, the centers nearby $c_l$ are likely candidates for the nearest center of $x$ in iteration $i+1$.
To speed up the assignment step, we thus only consider the $k_n$ nearest neighbours of $c_l$, $\mathcal{N}_{k_n}(c_l)$, as candidate centers for the points $x \in X_l$. 
Since for each point we only consider $k_n$ centers in the assignment step (in line~11 of Algorithm~\ref{alg:k2means}), the complexity is reduced to $O(n k_n d)$. In practice, we can set $k_n \ll k$.

We also use inequalities as in~\cite{elkan2003using} to avoid redundant distance computations in the assignment step (in line~11 of Algorithm~\ref{alg:k2means}). Note that we maintain only $nk_n$ lower bounds, for the neighbourhood of each point, instead of $nk$ for the Elkan method. We refer to the original Elkan paper~\cite{elkan2003using} for a detailed discussion on triangle inequalities and bounds.

As for standard Lloyd, the total energy can only decrease, both in the assignment step (since the points are moved to closer centers) and in the update step. Thus, the total energy is monotonically decreasing which guarantees convergence.

As shown by Arthur and Vassilvitskii~\cite{Arthur-DA-2007}, a good initialization, such as $k$-means++, often leads to a higher quality clustering compared to random sampling. Since the $O(ndk)$ complexity of $k$-means++ would negate the benefits of the $k^2$-means computation savings, we propose an alternative fast initialization scheme, which also leads to high quality clustering solutions.

\subsection{Greedy Divisive Initialization (GDI)}
For the initialization of our $k^2$-means, we propose a simple hierarchical clustering method named Greedy Divisive Initialization (GDI), detailed in Algorithm~\ref{alg:greedy_divisive_init}.
Similarly to other divisive clustering methods, such as~\cite{boley1998principal,su2004deterministic}, we start with a single cluster and repeatedly split the highest energy cluster until we reach $k$ clusters.

\begin{figure}
\begin{minipage}[c]{0.65\linewidth}
\centering
\begin{tabular}{c|c|c|c|}
\small
& Initialization & Iteration 1 & Iteration 2 \\ 
\hline
\parbox[b]{3mm}{\rotatebox[origin=l]{90}{\hspace{0.5cm} $k$-means}} & \includegraphics[width=0.25\textwidth]{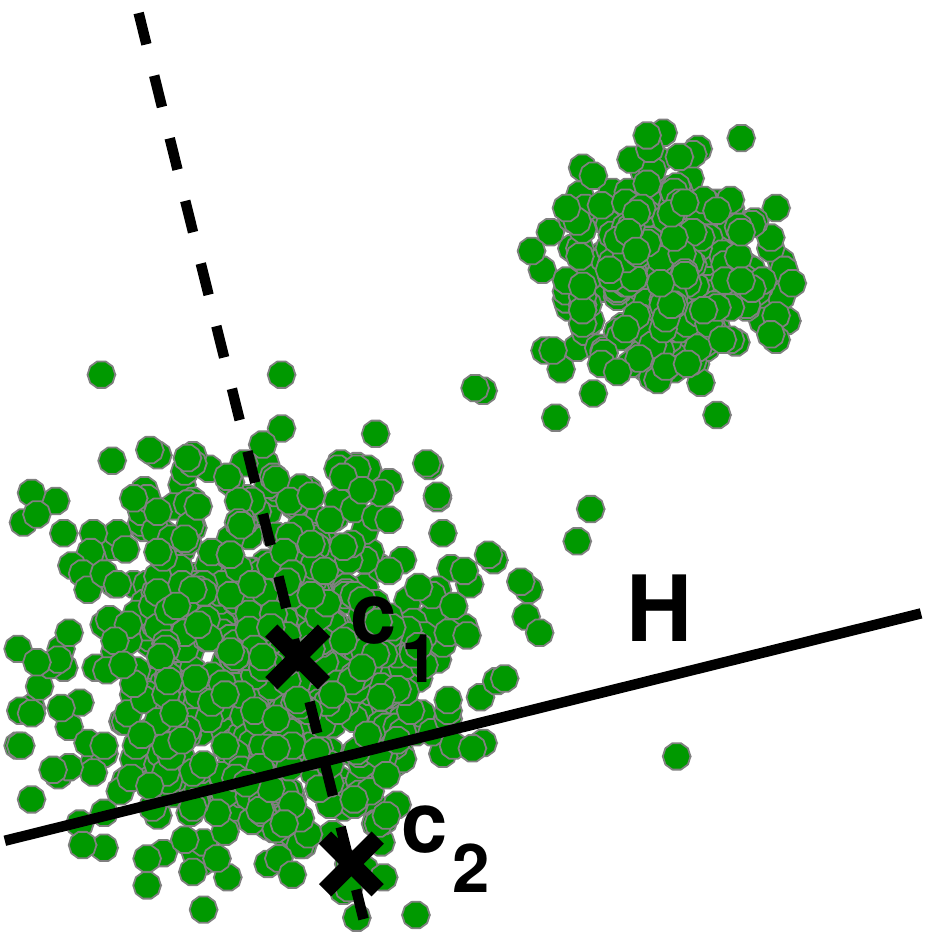} &
\includegraphics[width=0.25\textwidth]{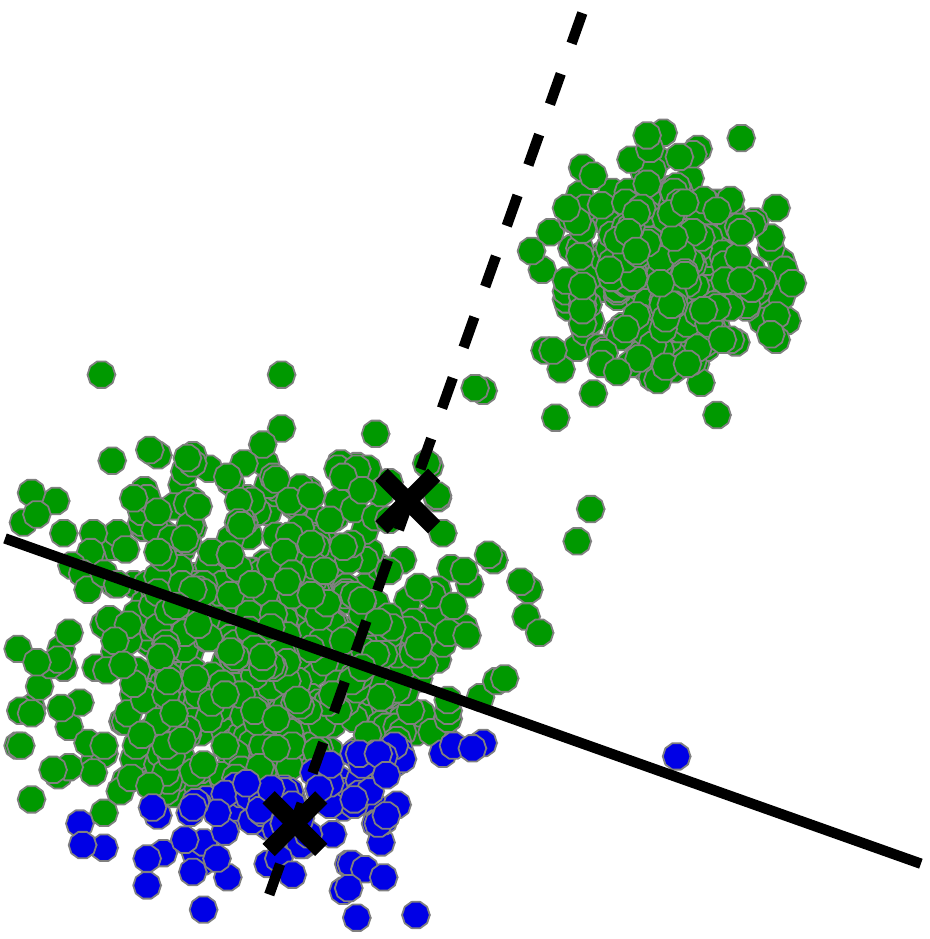} &
\includegraphics[width=0.25\textwidth]{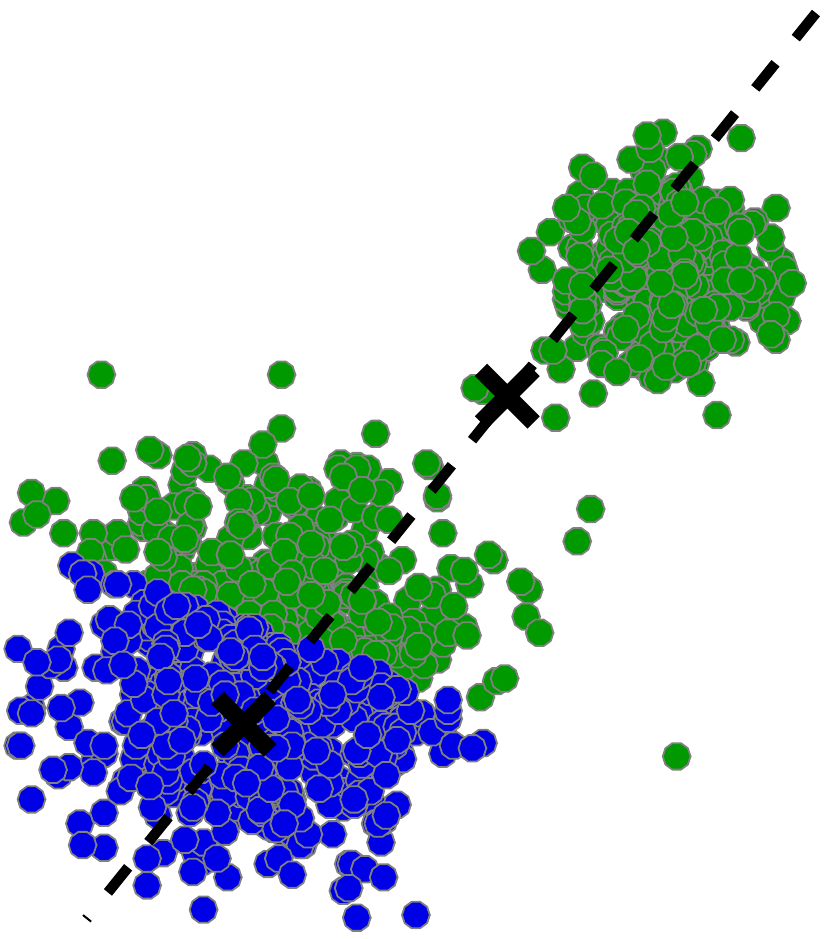} \\\hline
\parbox[b]{3mm}{\rotatebox[origin=l]{90}{\hspace{0.5cm} Projective Split}} &
\includegraphics[width=0.25\textwidth]{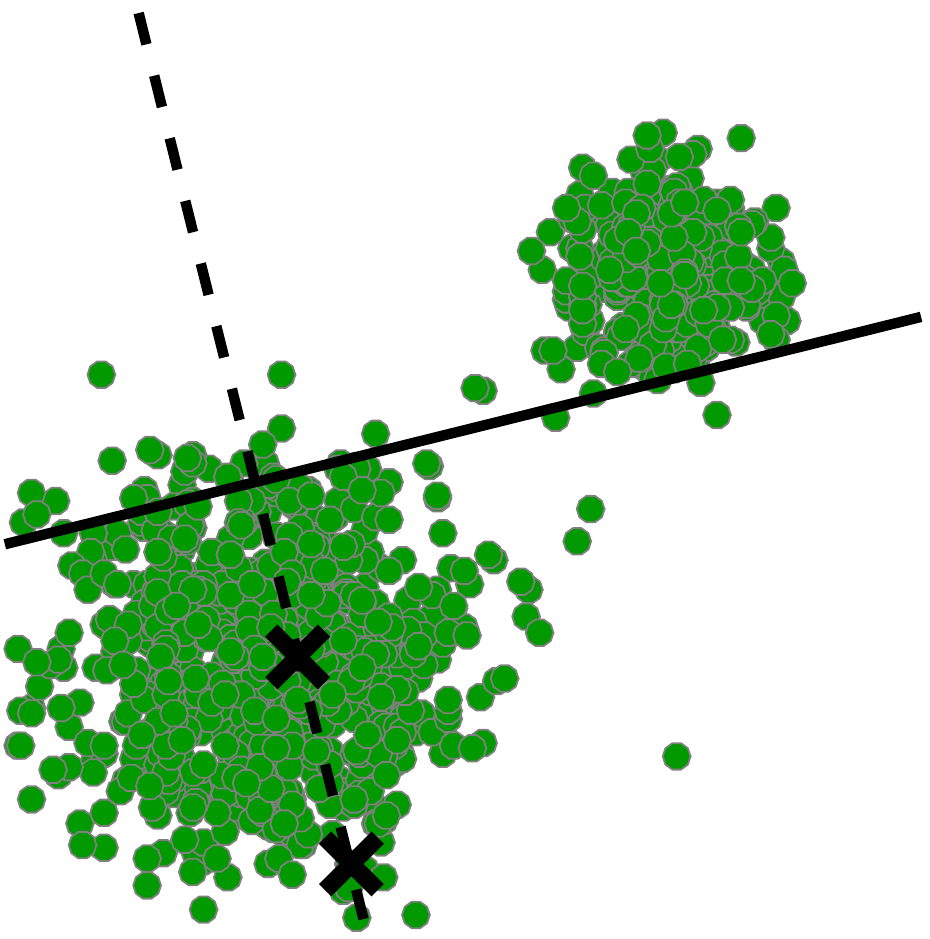} &
\includegraphics[width=0.25\textwidth]{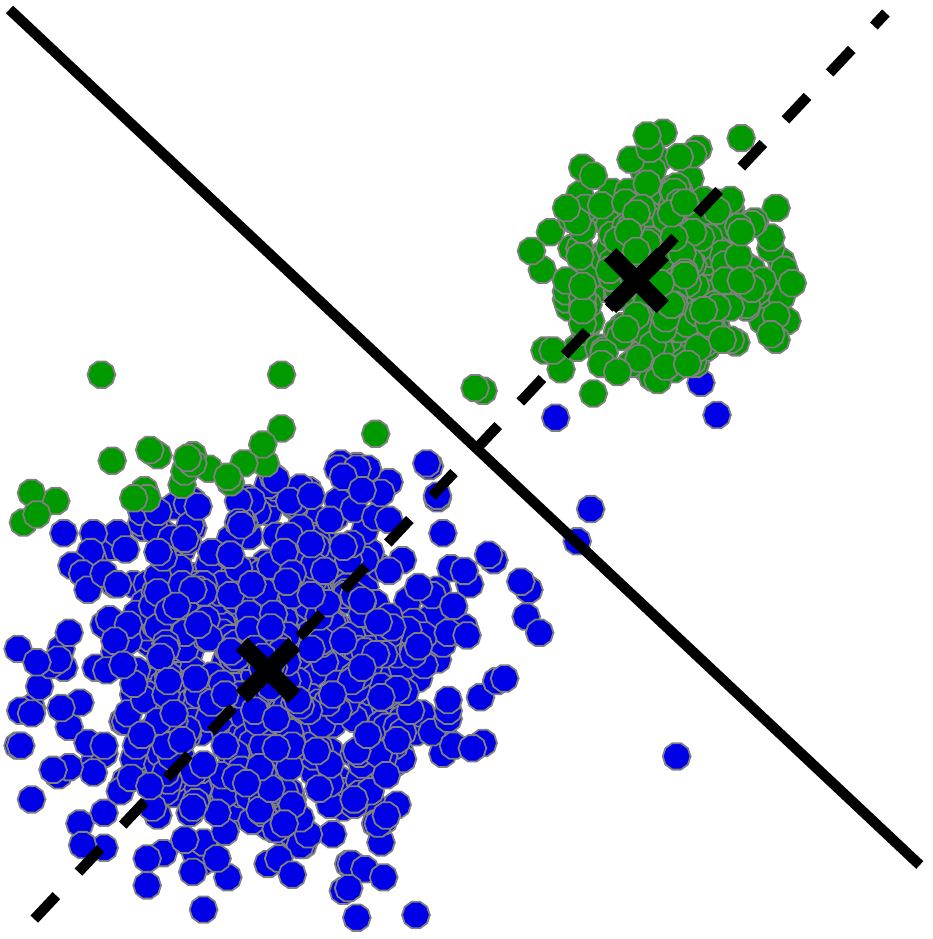} &
\includegraphics[width=0.25\textwidth]{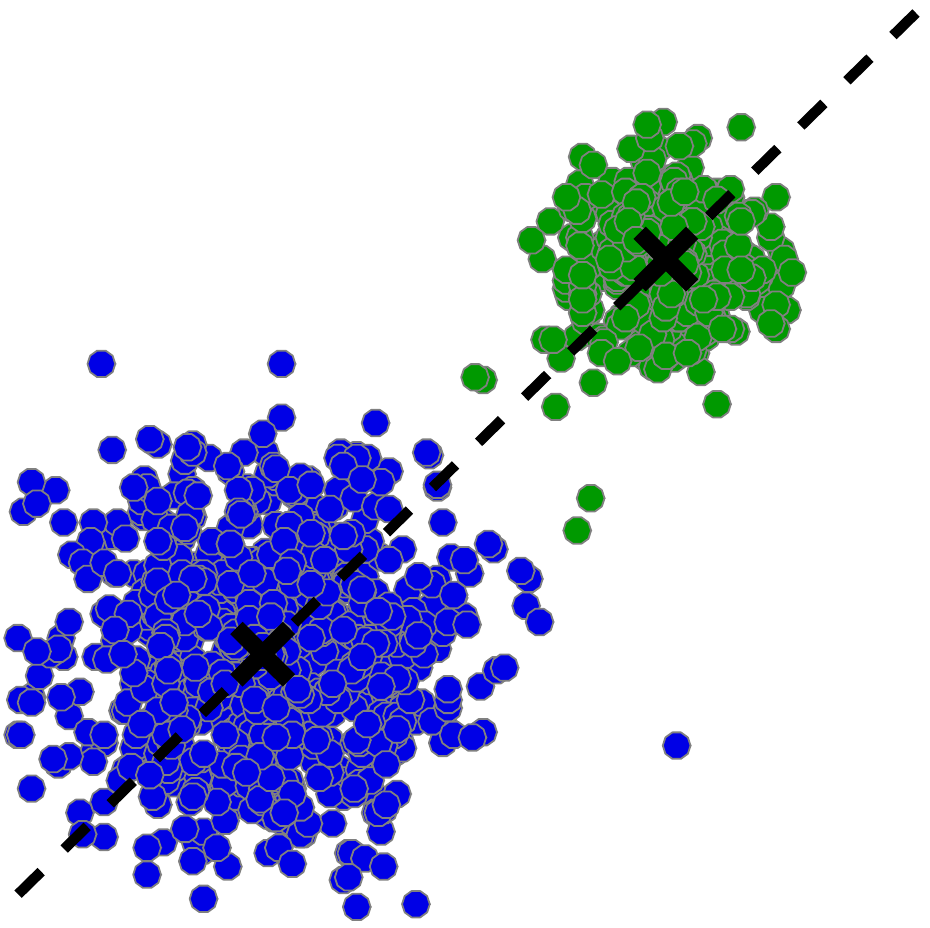} \\
\hline
\end{tabular}
\end{minipage}\hfill
\begin{minipage}[c]{0.34\linewidth}
\caption{Example of two iterations of Projective Split and standard $k$-means with $k=2$ using the same initialization. The dashed line shows the direction defined by two centers ($c_2-c_1$). The solid line shows where the algorithms split the data in each iteration. 
The splitting line of $k$-means always goes through the midpoint of the two centers, while Projective Split picks the minimal energy split along the dashed line. Even though the initial centers start in the same cluster, Projective Split can almost separate the clusters in a single iteration.}
\label{fig:projective_diagram}
\end{minipage}\hfill
\end{figure}

To efficiently split each cluster, we use Projective Split (Algorithm~\ref{alg:split_cluster}), a variant of $k$-means with $k=2$, that is motivated by the following observation: Suppose we have points $X'$ and centers $(c_1,c_2)$ in the $k$-means method. Let $H$ be the hyperplane with normal vector $c_2-c_1$, going through $\mu(c_1,c_2)$ (see e.g. the top left corner of Figure~\ref{fig:projective_diagram}). 
When we perform the standard $k$-means assignment step, we greedily assign each point to its closest centroid to get a solution with a lower energy, thus assigning the points on one side of $H$ to $c_1$, and the other side of $H$ to $c_2$.

Although this is the best assignment choice for the current centers $c_1$ and $c_2$, this may not be a good split of the data.
Therefore, we depart from the standard assignment step and consider instead \textit{all} hyperplanes along the direction $c_2-c_1$ (i.e. with normal vector $c_2-c_1$).
We project $X'$ onto $c_2-c_1$ and ``scan'' a hyperplane through the data to find the split that gives the lowest energy (lines~4-8 in Algorithm~\ref{alg:split_cluster}).
To efficiently recompute the energy of the cluster splits as the hyperplane is scanned, we use the following Lemma:
\begin{lemma}\label{energylemma}
\cite{kanungo2002local}[Lemma 2.1]
Let $S$ be a set of points with mean $\mu(S)$. Then for any point $z\in \R^d$
\begin{equation}
        \sum_{x \in S} \| x - z \|^2 = \sum_{x \in S} \| x - \mu(S) \|^2 + |S| \| z - \mu(S)  \|^2 
\end{equation}
\end{lemma}
We can now compute
\begin{align}
        &\phi(S \cup \{y\}) = \sum_{x \in S \cup \{y\}} \| x -  \mu (S \cup \{y\}) \|^2 \\ 
        &= \sum_{x \in S } \| x -  \mu (S \cup \{y\}) \|^2  
        + \| y -  \mu (S \cup \{y\}) \|^2 \label{lemmaused}\\
        &= \phi(S) + |S| \| \mu (S \cup \{y\}) - \mu(S) \|^2 + \| y -  \mu (S \cup \{y\}) \|^2  \label{incrementalupdate},
\end{align}
where we used Lemma~\ref{energylemma} in (\ref{lemmaused}).
Equipped with (\ref{incrementalupdate}) we can efficiently update energy terms in line~8 in Algorithm~\ref{alg:split_cluster} as we scan the hyperplane through the data $X_j$, using in total only $O(|X_j|)$ distance computations and mean updates. Note that $\mu(S\cup \{y\})$ is easily computed with an add operation as $( |S|\mu(S)+y )/(|S|+1)$.

\begin{minipage}[c]{0.48\linewidth}
\begin{algorithm}[H]
\caption{Greedy Divisive Initialization (GDI)}
\label{alg:greedy_divisive_init}
\begin{algorithmic}[1]
\small
\STATE \textbf{Given}: $k$, data $X$
\STATE Assign all points to one cluster
\STATE $C=\{\mu(X)\}$, $a(X)=1$ 
\WHILE{$|C|<k$}
\STATE Pick highest energy cluster:
\STATE $j \gets \argmax_{l} \phi( X_l )$
\STATE Split the cluster:
\STATE $X_a,c_a,X_b,c_b \gets \text{ProjectiveSplit}(X_j)$
\STATE $c_j \gets c_a$
\STATE $c_{|C|+1} \gets c_b$
\STATE $a(X_b) \gets |C|+1$
\STATE $C \gets C\cup \{c_{|C|+1}\}$
\ENDWHILE
\STATE \textbf{return} $C,a$
\end{algorithmic}
\end{algorithm}

\end{minipage}\hfill 
\begin{minipage}[c]{0.48\linewidth}
\begin{algorithm}[H]
\caption{Projective Split}
\label{alg:split_cluster}
\begin{algorithmic}[1]
\small
\STATE \textbf{Given}:data $X_j=(x_i)_{i=1}^{n_j}$ 
\STATE Pick two random samples $c_a, c_b$ from $X_j$
\WHILE{Not Converged}
    \STATE Sort $X_j$ along $c_a-c_b$: \label{proj_sort}
    \STATE $P_j \gets ( x_i \cdot (c_a-c_b) | x_i \in X_j )$
    \STATE $\tilde{X_j} \gets X_j \text{ sorted by } P_j$
    \STATE Find minimum-energy split:
    \STATE $l_{min} = \argmin_{l} \phi( (\tilde{x}_i)_{i=1}^l ) + \phi(  (\tilde{x}_i)_{i=l+1}^{n_j} )$
    \STATE $X_a \gets (\tilde{x}_i)_{i=1}^{l_{min}}$\label{optimalsplit}
    \STATE $X_b \gets (\tilde{x}_i)_{i=l_{min}+1}^{n_j}$
    \STATE $c_a, c_b \gets \mu(X_a), \mu(X_b)$
\ENDWHILE
\STATE \textbf{return} $X_a,c_a,X_b,c_b$
\end{algorithmic}
\end{algorithm}
\end{minipage}\hfill

Compared to standard $k$-means with $k=2$, our Projective Split takes the optimal split along the direction $c_2-c_1$ but greedily considers only this direction. 
In Figure~\ref{fig:projective_diagram} we show how this can lead to a faster convergence.

\subsection{Time Complexity}
Table~\ref{tab:complexity} shows the time and memory complexity of Lloyd, Elkan, MiniBatch, AKM, and our $k^2$-means.

The time complexity of each $k^2$-means iteration is dominated by two factors: building the nearest neighbour graph of $C$  (line~6), which costs $O(k^2)$ distance computations, as well as computing distances between points and candidate centers (line~11), which initially costs $nk_n$ distance computations.
Elkan and $k^2$-means use the triangle inequality to avoid redundant distance calculations and empirically we observe the $O(nkd)$ and $O(nk_n d)$ terms (respectively) gradually reduce down to $O(nd)$ at convergence.

In MiniBatch $k$-means processes only $b$ samples per iteration (with $b\ll n$) but needs more iterations for convergence. 
AKM limits the number of distance computations to $m$ per iteration, giving a complexity of $O(nmd)$.

Table~\ref{tab:initcomplexity} shows the time and memory complexity of random, $k$-means++ and our GDI initialization. 
For the GDI, the time complexity is dominated by calls to Projective Split.
If we limit Projective Split to maximum $O(1)$ iterations (2 in our experiments) then a call to ProjectiveSplit$(X_j)$ costs $O(|X_j|)$ distance computations and vector additions, $O(|X_j|)$ inner products and $O(|X_j|\log |X_j|)$ comparisons (for the sort), giving in total $O(|X_j|(\log |X_j|+d))$ complexity.
However, the resulting time complexity of GDI depends on the data.

For pathological datasets, it could happen for each call to ProjectiveSplit$(X')$, that the minimum split is of the form $\{y\},X'\setminus \{y\}$, i.e. only one point $y$ is split off.
In this case, for $|X|=n$, the total complexity will be $O(n(\log n +d) + (n-1)(\log(n-1) +d) + \cdots + (n-k)(\log(n-k) +d)) = O(nk(d+\log n))$.
\footnote{ 
A simple example of such a pathological dataset is $X=(x_i)_{i=1}^n \subset \R$ where $x_1=0$, $x_2 = 1$, $x_3=\phi(x_1,x_2)$, $x_4=\phi(x_1,x_2,x_3)$ and $x_n = \phi(x_1,\cdots,x_n)$. The size of $x_n$ grows extremely fast though, e.g. $x_{10}\approx 1581397605569$ and $x_{14}$ has 195 digits.}
    
A more reasonable case is when at each call ProjectiveSplit$(X')$ splits each cluster into two similarly large clusters, i.e. the minimum split is of the form $(X_a',X_b')$ where $|X_a|\approx|X_b|$. In this case the worst case scenario is when in each split the highest energy cluster is the largest cluster (in no. of samples), resulting a total complexity of $O(n\log k (d+\log n))$.
\footnote{If we split all clusters of approximately equal size simultaneously, we need $O(\log k)$ passes and perform $O(n (d+\log n))$ computations in each pass.}
Therefore the time complexity of GDI is somewhere between $O(n\log k (d + \log n))\sim O(n(d + \log n)k)$.

In our experiments we count vector operations for simplicity (i.e. dropping the $O(d)$ factor), as detailed in the next section. To fairly account for the $O(|X_j|\log |X_j|)$ complexity of the sorting step in ProjectiveSplit, we artificially count it as $|X_j|\log_2(|X_j|)/d$ vector operations.

\begin{table*}
\centering
\small
\tabcolsep=0.1cm
\resizebox{\linewidth}{!}{
\begin{tabular}{l|ccccc}
Complexity         & Lloyd  & Elkan~\cite{elkan2003using} & MiniBatch~\cite{Sculley-WWW-2010} & AKM~\cite{philbin2007object} & \textbf{$k^2$-means (ours)} \\\hline
 Time & $O(nkd)$  &  $O(nkd +k^2d) \sim O(nd +k^2d)$ & $O(bkd)$  &  $O(n m d)$  & $O(n k_n d + k^2d) \sim O(n d + k^2 d)$\\
 Memory & $O((n+k)d)$& $O((n+k)d + nk+k^2)$ & $O((b+k)d)$ & $O((n+k)d)$ & $O((n+k)d+nk_n+k^2)$\\ 
\end{tabular}
}
\vspace{-0.15cm}
\caption{Time and memory complexity per iteration for Lloyd, Elkan, MiniBatch, AKM and our $k^2$-means.}
\label{tab:complexity}
\end{table*}

\begin{table}
\centering
\small
\tabcolsep=0.15cm
\begin{tabular}{l|ccc}
Complexity & random & $k$-means++ & \textbf{GDI (ours)}  \\\hline
 Time    & $O(k)$ & $O(nkd)$     & $O(n (\log k)(d+\log n)) \sim O(nk(d+\log n))$ \\
 Memory  & $O(k)$ & $O(n+k)$     & $O(n+kd)$ \\ 
\end{tabular}
\vspace{-0.05cm}
\caption{Time and memory complexity for initialization. }
\label{tab:initcomplexity}
\vspace{-0.35cm}
\end{table}

\section{Experiments}
\label{sec:experiments}
For a fair comparison between methods implemented in various programming languages, we use the number of vector operations as a measure of complexity, i.e. distances, inner products and additions. While the operations all share an $O(d)$ complexity, the distance computations are most expensive accounting for the constant factor. However, since the runtime of all methods is dominated by distance computations (i.e. more than 95\% of the runtime), for simplicity we count all vector operations equally and refer to them as ``distance computations'', using the terminology from \cite{elkan2003using}.

\subsection{Datasets}
\label{ssc:datasets}

In our experiments we use datasets with 2414-150000 samples ranging from 50 to 32256  dimensions as listed in Table~\ref{tab:k2means_speedup10}. The datasets are diverse in content and feature representation.

To create \noindent{\textbf{cnnvoc}} we extract 4096-dimensional CNN features~\cite{krizhevsky2012imagenet} for 15662 bounding boxes, each belonging to 20 object categories, from PASCAL VOC 2007~\cite{everingham2010pascal} dataset. 
\noindent{\textbf{covtype}} uses the first 150000 entries of the Covertype dataset~\cite{uci-mlrepo} of cartographic features.
From the \textbf{mnist} database~\cite{lecun1998mnist} of handwritten digits we also generate \textbf{mnist50} by random projection of the raw pixels to a 50-dimensional subspace. For \textbf{tinygist10k} we use the first 10000 images with extracted gist features from the 80 million tiny images dataset~\cite{torralba200880}.  \noindent{\textbf{cifar}} represents 50000 training images from the CIFAR~\cite{krizhevsky2009learning} dataset.
\noindent{\textbf{usps}}~\cite{hull1994database} has scans of handwritten digits (raw pixels) from envelopes.
\noindent{\textbf{yale}} contains cropped face images from the Extended Yale B Database~\cite{GeBeKr01,KCLee05}. 

\subsection{Methods}
\label{ssc:methods}

We compare our \textbf{$k^2$-means} with relevant clustering methods: Lloyd (standard $k$-means), Elkan~\cite{elkan2003using} (accelerated Lloyd), MiniBatch~\cite{Sculley-WWW-2010} (web-scale online clustering), and AKM~\cite{philbin2007object} (efficient search structure). 

Aside from our {\bf GDI} initialization, we also use {\bf random} initialization and {\bf $k$-means++}~\cite{Arthur-DA-2007} in our experiments. For $k$-means++ we use the provided Matlab implementation.
We Matlab implement MiniBatch $k$-means according to Algorithm 1 in~\cite{Sculley-WWW-2010} and use the provided codes for Elkan and AKM.
{\bf Lloyd++} and {\bf Elkan++} combine $k$-means++ initialization with Lloyd and Elkan, respectively.

We run all methods, except MiniBatch, for a maximum of 100 iterations.
For MiniBatch $k$-means we use $b=100$ samples per batch and $t=n/2$ iterations.
For the Projective Split, Algorithm~\ref{alg:split_cluster}, we perform only 2 iterations.

\subsection{Initializations}
We compare $k$-means++, random and our GDI initialization by running 20 trials of $k$-means (Lloyd) clustering with $k\in \{100,200,500\}$ on the datasets (excluding cifar and tiny10k due to the prohibitive cost of standard Lloyd with a high number of clusters).
Table~\ref{tab:divisive_vs_others_greedy} reports minimum and average cluster energy as well as the average number of distance computations, relative to $k$-means++, averaged over the settings of $k$ (i.e. $20\times 3$ experiments for each row).

\begin{table*}
\centering
\small
\tabcolsep=0.2cm
\resizebox{0.95\linewidth}{!}
{
\begin{tabular}{l|ccc|ccc|cc}
&
\multicolumn{3}{c}{\textit{average convergence energy}} & 
\multicolumn{3}{|c|}{\textit{minimum convergence energy}} & 
\multicolumn{2}{c}{\textit{average runtime complexity}} \\
Dataset
& random & $k$-means++ & \textbf{GDI} 
& random & $k$-means++ & \textbf{GDI} 
& $k$-means++ &  \textbf{GDI} 
\\
\hline\hline
cnnvoc & 1.000 & 1.000 & 0.994 & 1.000 & 1.000 & 0.995 & 1.000 & 0.096 \\
covtype & 1.507 & 1.000 & 0.983 & 1.384 & 1.000 & 0.991 & 1.000 & 0.116 \\
mnist & 1.000 & 1.000 & 0.999 & 1.000 & 1.000 & 0.999 & 1.000 & 0.093 \\
mnist50 & 1.000 & 1.000 & 0.999 & 1.000 & 1.000 & 0.999 & 1.000 & 0.119 \\
tinygist10k & 0.999 & 1.000 & 0.994 & 1.000 & 1.000 & 0.996 & 1.000 & 0.098 \\
usps & 1.019 & 1.000 & 0.996 & 1.016 & 1.000 & 0.996 & 1.000 & 0.099 \\
yale & 1.024 & 1.000 & 1.008 & 1.025 & 1.000 & 1.005 & 1.000 & 0.103 \\\hline
\textbf{average} & 1.078 & 1.000 & 0.996 & 1.061 & 1.000 & 0.997 & 1.000 & 0.103 \\

\end{tabular}
}
\vspace{-0.2cm}
\caption{Comparison of energy and runtime complexity for random, $k$-means++, and our GDI initialization. The results are displayed relative to $k$-means++, averaged over $20\times 3$ configurations. Random initialization does not require distance computations.}
\label{tab:divisive_vs_others_greedy}
\vspace{-0.3cm}
\end{table*}

Our GDI gives a (slightly) better average and minimum convergence energy than the other initializations, while its runtime complexity  is an order of magnitude smaller than in the case of $k$-means++ initialization.

The corresponding expanded Table~7 of the supplementary materials shows that 
 speedup of GDI over $k$-means++ improves as $k$ grows, and at $k=500$ is typically more than an order of magnitude.
This makes GDI a good choice for the initialization of $k^2$-means.

\subsection{Performance} 
Our goal is \textit{fast accurate clustering}, where the cluster energy differs only slightly from Lloyd with a good initialization (such as $k$-means++) at convergence. 
Therefore, we measure the runtime complexity needed to achieve a clustering energy that is within $1\%$ of the energy obtained with Lloyd++ at convergence. 
In the supplementary material we report on performance for more reference levels ($0\%,0.5\%$ and $2\%$).

For a given budget i.e. the maximum number of iterations and parameters such as $m$ for AKM and $k_n$ for $k^2$ means, it is not known beforehand how well the algorithms approximate the targeted Lloyd++ energy. 
For a fair comparison we use an oracle to select the best parameters and the  number of iterations for each method, i.e. the ones that give the highest speedup but still reach the reference error. 
In practice, one can use a rule of thumb or progressively increase $k$, $m$ and the number of iterations until a desired energy has been reached.

\begin{table}[t!]
\vspace{-0.1cm}
\centering
  \begin{minipage}[c]{0.48\linewidth}
  \centering
\small
\tabcolsep=0.1cm
\resizebox{0.96\linewidth}{!}
{
\begin{tabular}{l|c|cccccc}
Dataset & $k$
& AKM & Elkan++ & Elkan & Lloyd++ & Lloyd  & \textbf{$k^2$-means}
\\
\hline\hline
cifar & 50 & 1.0 & 2.6 & 3.7 & 1.0 & 1.0 & \textbf{9.5} \\
$n=50000 $ & 200 & 1.9 & 3.0 & 4.6 & 1.0 & 1.1 & \textbf{26.2} \\
$d=3072 $ & 1000 & 4.9 & 3.0 & 5.1 & 1.0 & 1.2 & \textbf{86.7} \\
cnnvoc & 50 & \textbf{13.8} & 2.1 & 2.9 & 1.0 & 1.4 & 9.0 \\
$n=15662 $ & 200 & \textbf{22.6} & 2.0 & 2.8 & 1.0 & 1.2 & 19.2 \\
$d=4096 $  & 1000 & 3.3 & 1.9 & 2.8 & 1.0 & 0.9 & \textbf{20.2} \\
covtype & 50 & - & 6.1 & - & 1.0 & - & \textbf{35.1} \\
$n=150000 $ & 200 & - & 6.3 & - & 1.0 & - & \textbf{78.7} \\
$d=54 $  & 1000 & - & 8.5 & - & 1.0 & - & \textbf{176.6} \\
mnist & 50 & 7.3 & 3.6 & 5.3 & 1.0 & 1.5 & \textbf{12.3} \\
$n=60000 $ & 200 & 1.9 & 3.7 & 5.7 & 1.0 & 1.2 & \textbf{24.6} \\
$d=784 $  & 1000 & 4.7 & 3.6 & 5.9 & 1.0 & 0.8 & \textbf{43.4} \\
mnist50 & 50 & \textbf{12.7} & 3.7 & 5.4 & 1.0 & 1.3 & 8.8 \\
$n=60000 $ & 200 & 1.9 & 4.2 & 6.7 & 1.0 & 1.2 & \textbf{22.3} \\
$d=50 $  & 1000 & 3.1 & 4.1 & 6.6 & 1.0 & 0.8 & \textbf{38.0} \\
tinygist10k & 50 & \textbf{16.2} & 2.4 & 3.6 & 1.0 & 1.4 & 11.7 \\
$n=10000 $ & 200 & 12.8 & 2.3 & 3.5 & 1.0 & 1.3 & \textbf{22.3} \\
$d=384 $  & 1000 & 1.5 & 2.1 & - & 1.0 & - & \textbf{13.6} \\
usps & 50 & 5.3 & 4.1 & - & 1.0 & - & \textbf{11.8} \\
$n=7291 $ & 200 & 16.8 & 4.4 & - & 1.0 & - & \textbf{23.6} \\
$d=256 $ & 1000 & \textbf{18.5} & 2.7 & - & 1.0 & - & - \\
yale & 50 & 2.1 & 4.2 & 6.3 & 1.0 & 0.6 & \textbf{17.9} \\
$n=2414 $ & 200 & \textbf{21.9} & 2.9 & - & 1.0 & - & 13.9 \\
$d=32256 $  & 1000 & - & \textbf{1.9} & - & 1.0 & - & - \\\hline
avg. speedup & & 8.7 &  3.6 & 4.7 & 1.0 & 1.1 & \textbf{33.0}\\

\end{tabular}
}
\vspace{0.05cm}
\caption{Algorithmic speedup in reaching an energy within $1\%$ from the final Lloyd++ energy. (-) marks failure in reaching the target of $1\%$ relative error.
For each method, the parameter(s) that gave the highest speedup at 1\% error is used. 
}
\label{tab:k2means_speedup10}
   \end{minipage}\hfill 
\begin{minipage}[c]{0.48\linewidth}
\centering
\small
\tabcolsep=0.1cm
\resizebox{0.96\linewidth}{!}
{
\begin{tabular}{l|c|cccccc}
Dataset & $k$
& AKM & Elkan++ & Elkan & Lloyd++ & Lloyd  & \textbf{$k^2$-means}
\\
\hline\hline
cifar & 50 & - & 17.8 & - & 1.0 & - &  \textbf{37.9} \\
& 200 & 1.2 & 24.2 & - & 1.0 & - &  \textbf{139.8} \\
& 1000 & 11.3 & 17.5 & 28.2 & 1.0 & 2.6 &  \textbf{373.6} \\
cnnvoc & 50 & 2.4 & 9.3 & - & 1.0 & - &  \textbf{26.2} \\
& 200 & 3.7 & 9.3 & - & 1.0 & - &  \textbf{59.7} \\
& 1000 & 5.8 & \textbf{8.1} & - & 1.0 & - &  - \\
covtype & 50 & - & 28.9 & - & 1.0 & - &  \textbf{172.0} \\
& 200 & - & 40.2 & - & 1.0 & - &  \textbf{442.4} \\
& 1000 & - & \textbf{44.5} & - & 1.0 & - &  - \\
mnist & 50 & 1.1 & 17.3 & 26.6 & 1.0 & 2.9 &  \textbf{39.3} \\
& 200 & - & 25.8 & - & 1.0 & - &  \textbf{81.0} \\
& 1000 & 9.0 & 29.8 & - & 1.0 & - &  \textbf{141.1} \\
mnist50 & 50 & - & 18.7 & - & 1.0 & - &  \textbf{31.0} \\
& 200 & 2.1 & 26.6 & - & 1.0 & - &  \textbf{80.3} \\
& 1000 & 5.1 & 22.7 & - & 1.0 & - &  \textbf{94.1} \\
tinygist10k & 50 & 12.5 & 12.5 & 20.1 & 1.0 & 3.6 &  \textbf{50.1} \\
& 200 & 4.7 & 11.0 & - & 1.0 & - &  \textbf{71.8} \\
& 1000 & 2.6 & \textbf{7.8} & - & 1.0 & - &  - \\
usps & 50 & - & 12.6 & - & 1.0 & - &  \textbf{31.7} \\
& 200 & 3.4 & 14.6 & - & 1.0 & - &  \textbf{54.4} \\
& 1000 & - & \textbf{9.4} & - & 1.0 & - &  - \\
yale & 50 & 2.8 & 9.5 & - & 1.0 & - &  \textbf{32.5} \\
& 200 & \textbf{20.8} & 6.5 & - & 1.0 &  - & 18.7 \\
& 1000 & - & \textbf{4.0} & - & 1.0 & - &  - \\\hline
avg. speedup & & 5.9 &  17.9 & 25.0 & 1.0 & 3.0 & \textbf{104.1}\\

\end{tabular}
}
\vspace{0.05cm}
\caption{Algorithmic speedup in reaching the same energy as the final Lloyd++ energy. (-) marks failure in reaching the target of $0\%$ relative error.
For each method, the parameter(s) that gave the highest speedup at 0\% error is used.  }
\label{tab:k2means_speedup00}
  \end{minipage}\hfill  
\vspace{-0.35cm}
\end{table}

To measure performance we run AKM, Elkan++, Elkan, Lloyd++, Lloyd, MiniBatch, and $k^2$-means with $k\in\{50,200,1000\}$ on various datasets, with 3 different seeds and report average speedups over Lloyd++ when the energy reached is within $1\%$ from Lloyd++ at convergence in Table~\ref{tab:k2means_speedup10}.

Each method is stopped once it reaches the reference energy and for AKM and $k^2$-means, we use the parameters $m$ and $k_n$ from $\{3,5,10,20,30,50,100,200\}$ that give the highest speedup.

Table~\ref{tab:k2means_speedup10} shows that for most settings, our $k^2$-means has the highest algorithmic speedup at $1\%$ error. 
It benefits the most when both the number of clusters and the number of points are large, e.g. for $k=200$ at least $19\times$ speedup for all datasets with $n\ge 7000$ samples. 
We do not reach the target energy for usps and yale with  $k=1000$, because $k_n$ was limited to $200$.

\begin{figure}
\centering
\resizebox{\linewidth}{!}{
\begin{tabular}{cc}
\renewcommand*{\arraystretch}{0.1}
\includegraphics[width=0.32\textwidth]{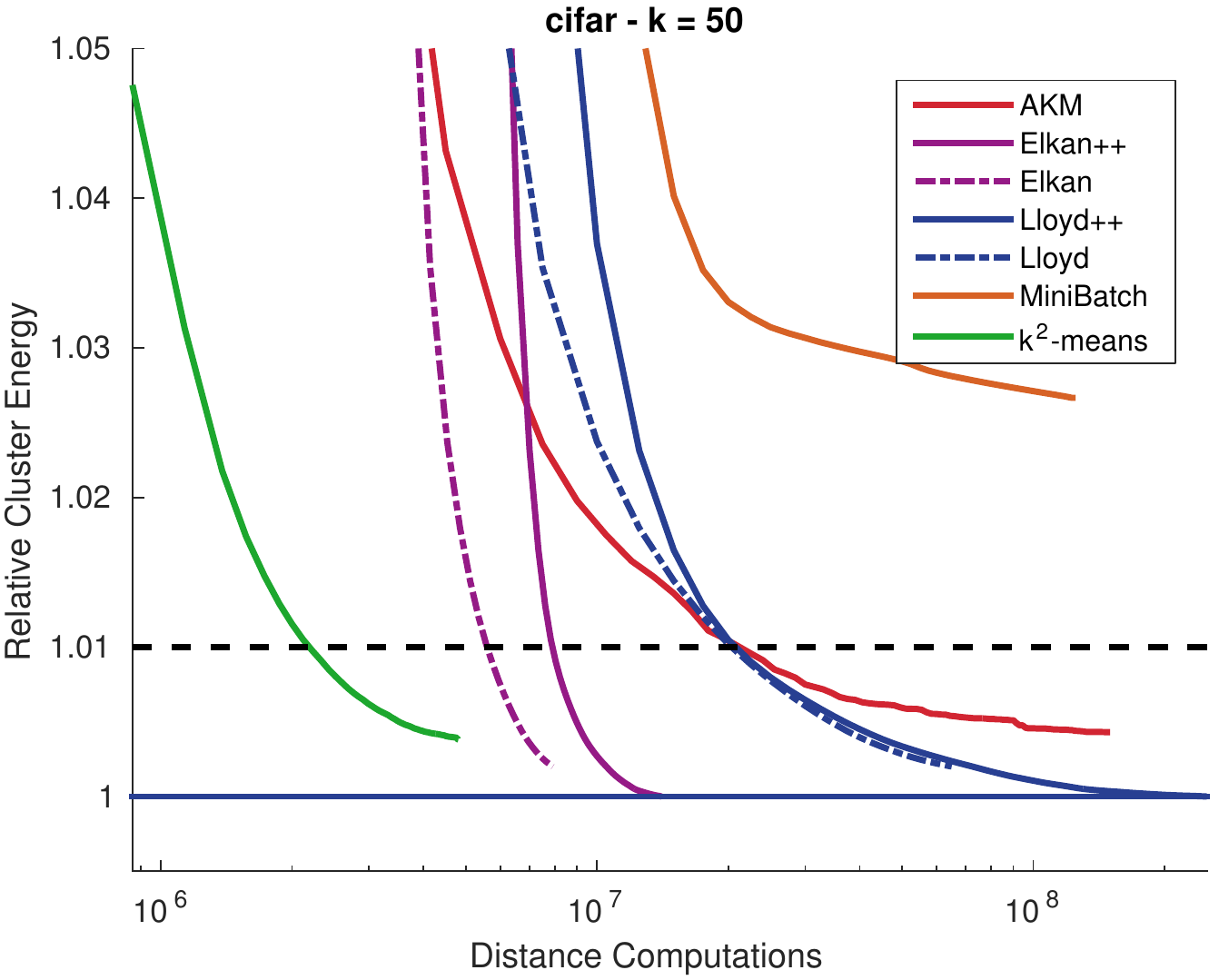} &
\includegraphics[width=0.32\textwidth]{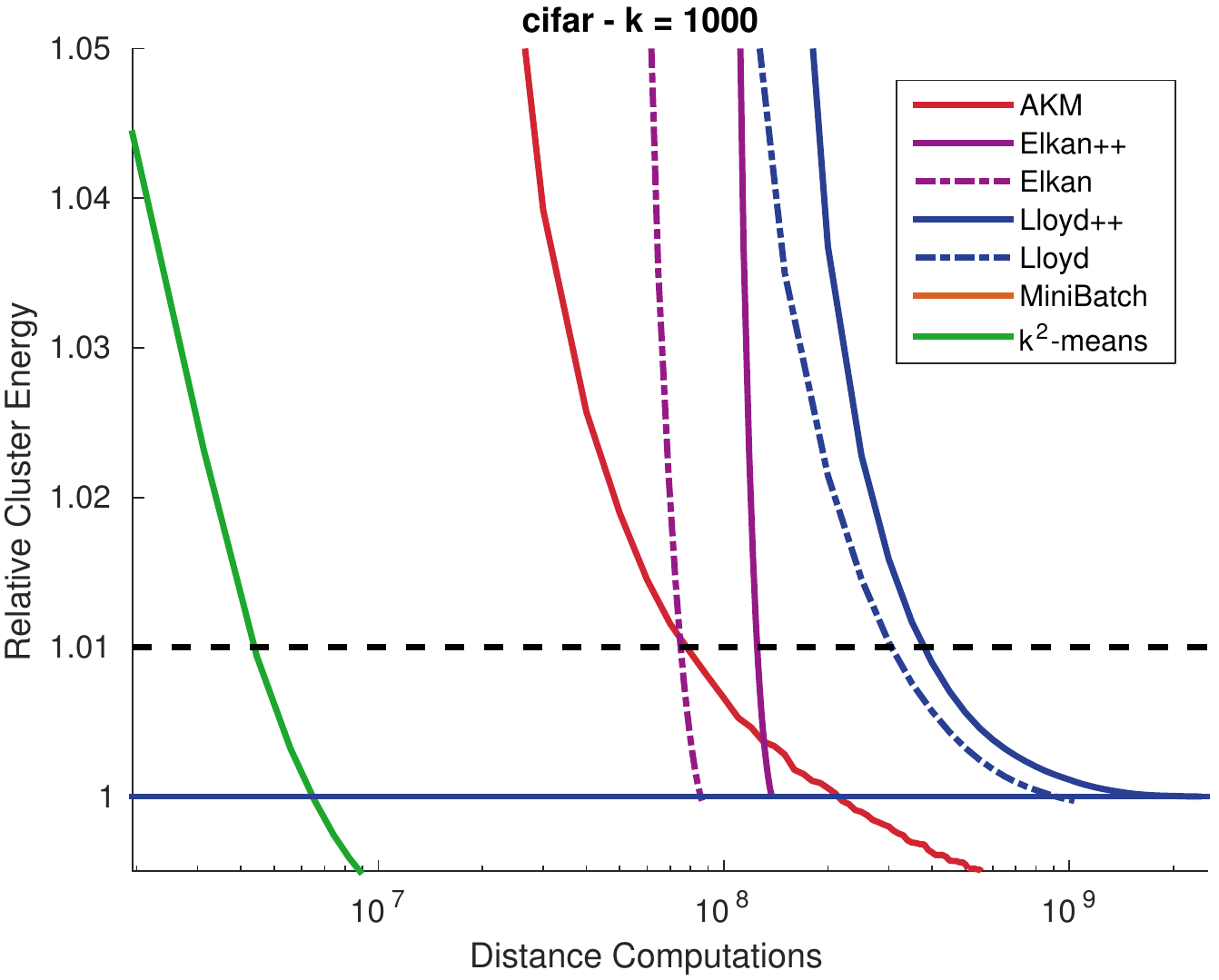} \\
\includegraphics[width=0.32\textwidth]{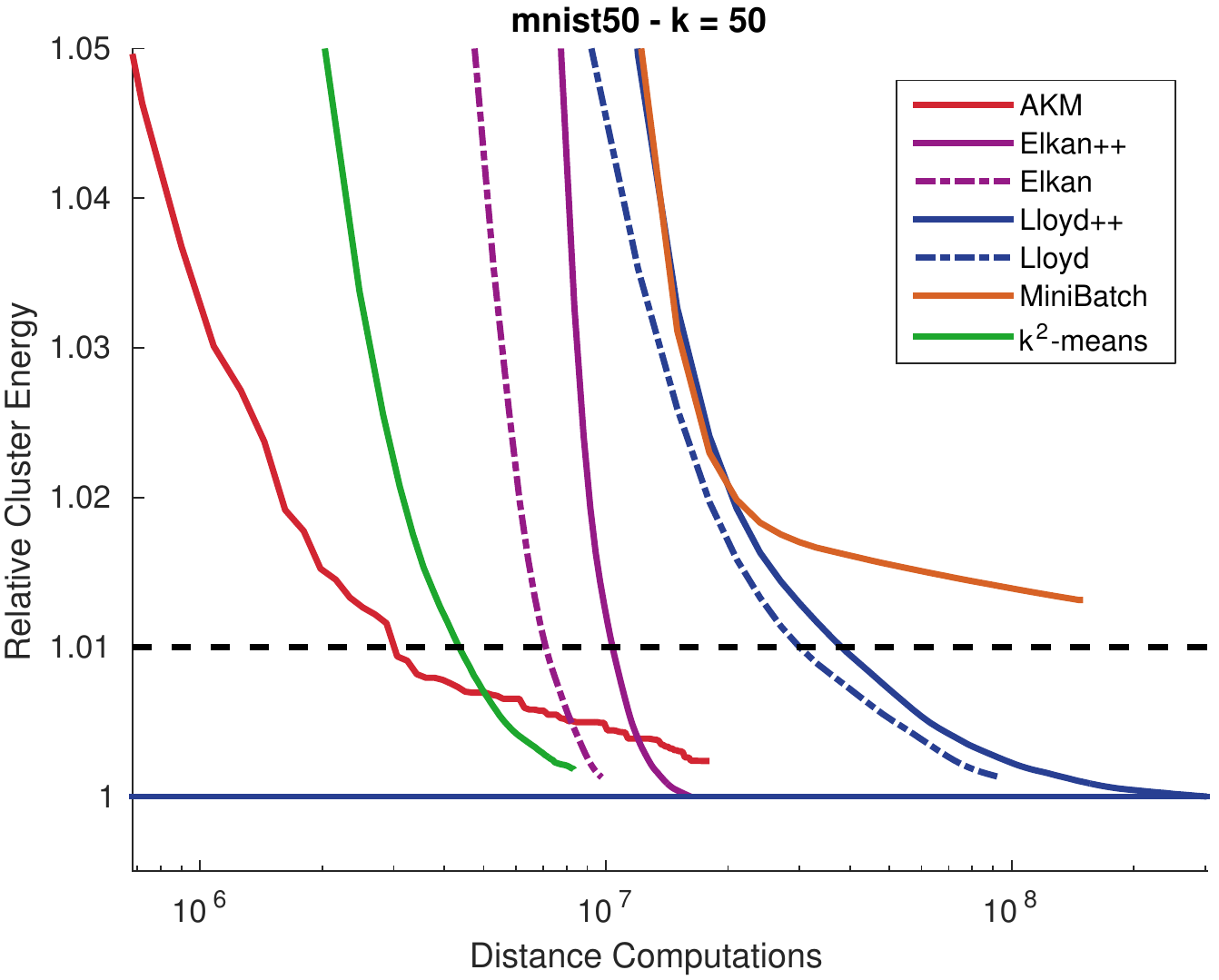} &
\includegraphics[width=0.32\textwidth]{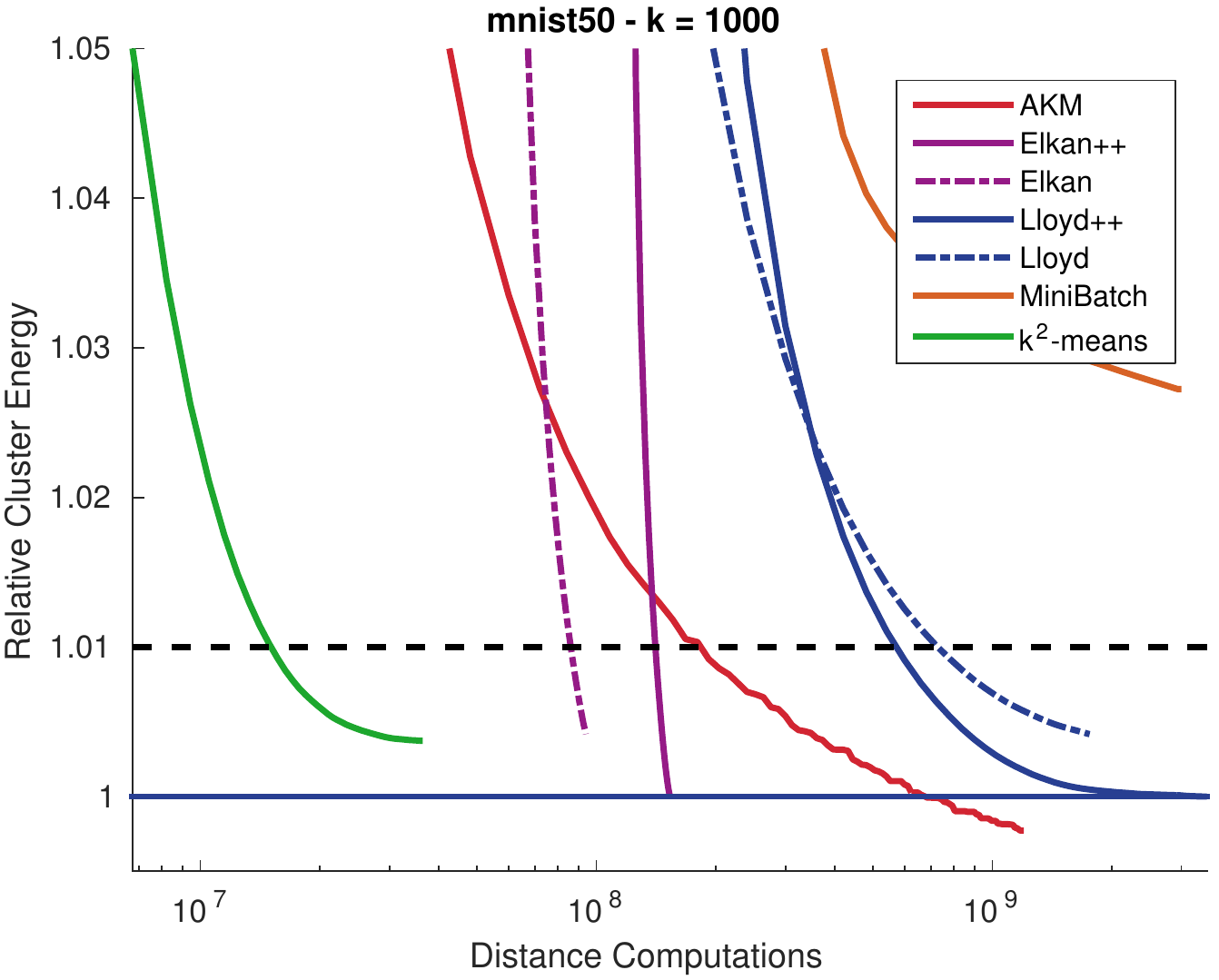} \\
\end{tabular}
}
\vspace{-0.2cm}
\caption{Cluster Energy (relative to best Lloyd++ energy) vs distance computations on cifar and mnist50 for $k\in\{50,1000\}$. For AKM and $k^2$-means, we use the parameter with the highest algorithmic speedup at $1\%$ error. }
\label{fig:k2means_plot}
\end{figure}

Figure~\ref{fig:k2means_plot} show the convergence curves corresponding to cifar and mnist50 entries in Table~\ref{tab:k2means_speedup10}. 
On cifar the benefit of $k^2$-means is clear since it reaches the reference error significantly faster than the other methods. 
On mnist50 $k^2$-means is considerably faster than AKM for $k=1000$ but AKM reaches the $1\%$ reference faster for $k=50$. We show more convergence curves in the supplementary material.

In all settings of Table~\ref{tab:k2means_speedup10}, Elkan++ gives a consistent up to $8.5\times$ speedup (since it is an exact acceleration of Lloyd++). For some settings Elkan is faster than Elkan++ in reaching the desired accuracy. This is due to the faster initialization.
MiniBatch fails in all but one case (mnist, $k=50$) to reach the reference error of $1\%$ and is thus not shown.

For accurate clustering, when the reference energy is the Lloyd++ convergence energy (i.e. 0\% error), Table~\ref{tab:k2means_speedup00} shows that the speedups of $k^2$-means are even higher. For this setting, the second fastest method is Elkan++, which is designed for accelerating the exact Lloyd++.

In the \textbf{supplementary material} we report extra results at more energy reference levels and also more plots showing the convergence of the compared methods on different datasets.

\section{Conclusions}
\label{sec:conclusions}
We proposed $k^2$-means, a simple yet efficient method ideally suited for fast and accurate large scale clustering ($n>10000$, $k>100$, $d>50$).
$k^2$-means combines an efficient divisive initialization with a new method to speed up the $k$-means iterations by using the $k_n$ nearest clusters as the new set of candidate centers for the cluster members as well as triangle inequalities.
The algorithmic complexity of our $k^2$-means is sublinear in $k$ for $n\gg k$ and experimentally shown  to give a high accuracy on diverse datasets.
For accurate clustering, $k^2$-means requires an order of magnitude fewer computations than alternative methods such as the fast approximate $k$-means (AKM) clustering.
Moreover, our efficient divisive initialization leads to comparable clustering energies and significantly lower runtimes than the $k$-means++ initialization under the same conditions.

\small
\bibliography{k2means}
\bibliographystyle{abbrv}

\pagebreak
{
\setcounter{page}{1}
\widetext
\begin{center}
\textbf{\large k$^2$-means for fast and accurate large scale clustering\\--supplementary material--}
\end{center}
\begin{abstract} 
This document is the supplementary material of our paper, providing further experimental results.
We show a more detailed comparison of the initializations, convergence curves for more datasets and settings as well as
a more detailed speedup table when the reference energy is taken at 0\%, 0.5\%, 1\% and 2\% deviation from the Lloyd++ convergence energy.
\end{abstract}

\section*{Additional experimental results}

In  the expanded Table~\ref{tab:divisive_vs_others_greedy_expand}, corresponding to Table~4 in the paper, we see that the  speedup of GDI over $k$-means++ improves as $k$ grows, and at $k=500$ is typically more than an order of magnitude.

In Figure~\ref{fig:k2means_plot} we show the convergence curves corresponding to cifar, cnnvoc, mnist and mnist50 for $k\in\{50,200,1000\}$, which correspond to entries in Table~5 in the paper. For the same datasets we further show in Figure~\ref{fig:k2means_vs_ann_plot} the convergence curves of all parameters tried for $k^2$-means and AKM. Note that we do not try parameters $k_n$ and $m$ that are larger than $k$.

In Tables~\ref{tab:k2means_speedup00_exp},\ref{tab:k2means_speedup5_exp},\ref{tab:k2means_speedup10_exp} and \ref{tab:k2means_speedup20_exp} we  show a detailed speedup table when the reference energy is taken at 0\%, 0.5\%, 1\% and 2\% deviation from the Lloyd++ convergence energy.
The algorithmic speedups of our method are the largest when the aim is to reach the same energy as the final Lloyd++ energy (i.e. 0\% error). 
However, for a less accurate result, i.e. when the reference is taken at 2\% from Lloyd++, Table \ref{tab:k2means_speedup20_exp} shows that $k^2$-means is still competitive with AKM.

We conclude that our $k^2$-means reaches its full potential when used for \textit{accurate} clustering, that is, low clustering energies comparable with the energies at the convergence of a standard Lloyd with a robust $k$-means++ initialization.
In such conditions, $k^2$-means is clearly orders of magnitude faster than Lloyd++ and significantly faster than the efficient AKM and Elkan++.

\begin{table*}[h]
\centering
\small
\tabcolsep=0.1cm
\resizebox{0.95\linewidth}{!}
{
\begin{tabular}{lc|ccc|ccc|cc}
& &
\multicolumn{3}{c}{\textit{average convergence energy}} & 
\multicolumn{3}{|c|}{\textit{minimum convergence energy}} & 
\multicolumn{2}{c}{\textit{average runtime complexity}} \\
Dataset & k
& random & $k$-means++ & \textbf{GDI} 
& random & $k$-means++ & \textbf{GDI} 
& $k$-means++ &  \textbf{GDI} 
\\
\hline\hline
cnnvoc & 100 & \textbf{1.00} & \textbf{1.00} & \textbf{1.00} & \textbf{1.00} & \textbf{1.00} & \textbf{1.00} & 1.00 & \textbf{0.16} \\
& 200 & 1.00 & 1.00 & \textbf{0.99} & 1.00 & 1.00 & \textbf{0.99} & 1.00 & \textbf{0.09} \\& 500 & 1.00 & 1.00 & \textbf{0.99} & 1.00 & 1.00 & \textbf{0.99} & 1.00 & \textbf{0.04} \\covtype & 100 & 1.51 & 1.00 & \textbf{0.99} & 1.47 & 1.00 & \textbf{0.99} & 1.00 & \textbf{0.19} \\
& 200 & 1.58 & 1.00 & \textbf{0.98} & 1.38 & 1.00 & \textbf{0.99} & 1.00 & \textbf{0.11} \\& 500 & 1.43 & 1.00 & \textbf{0.99} & 1.30 & 1.00 & \textbf{0.99} & 1.00 & \textbf{0.05} \\mnist & 100 & \textbf{1.00} & \textbf{1.00} & \textbf{1.00} & \textbf{1.00} & \textbf{1.00} & \textbf{1.00} & 1.00 & \textbf{0.15} \\
& 200 & \textbf{1.00} & \textbf{1.00} & \textbf{1.00} & \textbf{1.00} & \textbf{1.00} & \textbf{1.00} & 1.00 & \textbf{0.09} \\& 500 & \textbf{1.00} & \textbf{1.00} & \textbf{1.00} & \textbf{1.00} & \textbf{1.00} & \textbf{1.00} & 1.00 & \textbf{0.04} \\mnist50 & 100 & \textbf{1.00} & \textbf{1.00} & \textbf{1.00} & \textbf{1.00} & \textbf{1.00} & \textbf{1.00} & 1.00 & \textbf{0.19} \\
& 200 & \textbf{1.00} & \textbf{1.00} & \textbf{1.00} & \textbf{1.00} & \textbf{1.00} & \textbf{1.00} & 1.00 & \textbf{0.11} \\& 500 & \textbf{1.00} & \textbf{1.00} & \textbf{1.00} & \textbf{1.00} & \textbf{1.00} & \textbf{1.00} & 1.00 & \textbf{0.05} \\tinygist10k & 100 & \textbf{1.00} & \textbf{1.00} & \textbf{1.00} & \textbf{1.00} & \textbf{1.00} & \textbf{1.00} & 1.00 & \textbf{0.16} \\
& 200 & 1.00 & 1.00 & \textbf{0.99} & \textbf{1.00} & \textbf{1.00} & \textbf{1.00} & 1.00 & \textbf{0.09} \\& 500 & 1.00 & 1.00 & \textbf{0.99} & \textbf{1.00} & \textbf{1.00} & \textbf{1.00} & 1.00 & \textbf{0.04} \\usps & 100 & 1.01 & 1.00 & \textbf{0.99} & 1.01 & \textbf{1.00} & \textbf{1.00} & 1.00 & \textbf{0.16} \\
& 200 & 1.01 & 1.00 & \textbf{0.99} & 1.01 & 1.00 & \textbf{0.99} & 1.00 & \textbf{0.09} \\& 500 & 1.04 & \textbf{1.00} & \textbf{1.00} & 1.04 & \textbf{1.00} & \textbf{1.00} & 1.00 & \textbf{0.05} \\yale & 100 & 1.01 & \textbf{1.00} & \textbf{1.00} & 1.00 & 1.00 & \textbf{0.99} & 1.00 & \textbf{0.16} \\
& 200 & 1.02 & \textbf{1.00} & \textbf{1.00} & 1.02 & \textbf{1.00} & \textbf{1.00} & 1.00 & \textbf{0.10} \\& 500 & 1.05 & \textbf{1.00} & 1.03 & 1.05 & \textbf{1.00} & 1.02 & 1.00 & \textbf{0.05} \\
\end{tabular}
}
\vspace{-0.2cm}
\caption{Comparison of energy and runtime complexity for random, $k$-means++, and our GDI initialization. The results are displayed relative to $k$-means++, averaged over $20$ seeds. Random initialization does not require distance computations.}
\label{tab:divisive_vs_others_greedy_expand}
\vspace{-0.3cm}
\end{table*}

\begin{figure*}
\centering
\resizebox{\linewidth}{!}{
\begin{tabular}{ccc}
\includegraphics[width=0.32\textwidth]{cluster_all_k2elkan_philbin_plusplus_cifar_ce_dists_k=50.pdf} &
\includegraphics[width=0.32\textwidth]{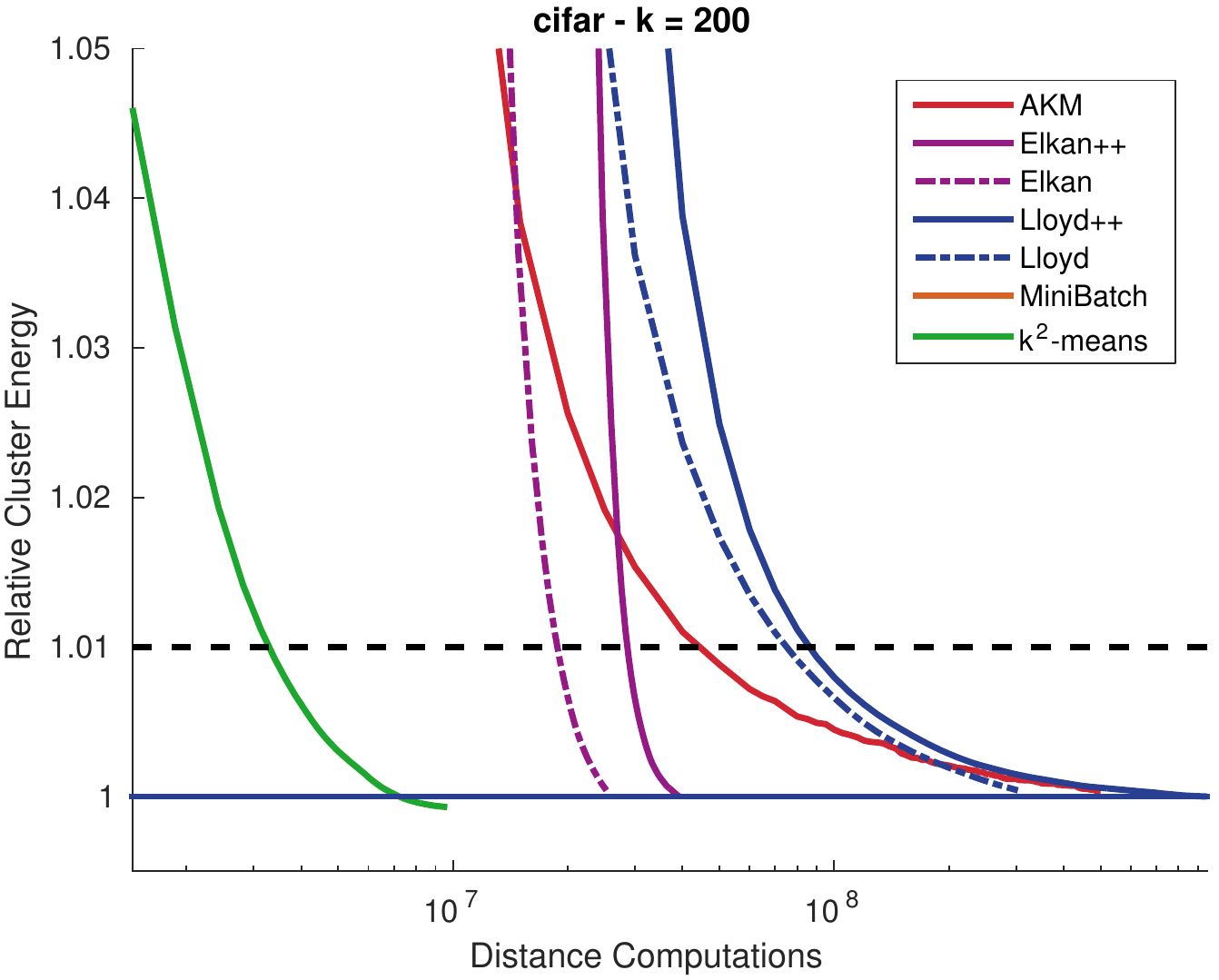} &
\includegraphics[width=0.32\textwidth]{cluster_all_k2elkan_philbin_plusplus_cifar_ce_dists_k=1000.pdf} \\
\includegraphics[width=0.32\textwidth]{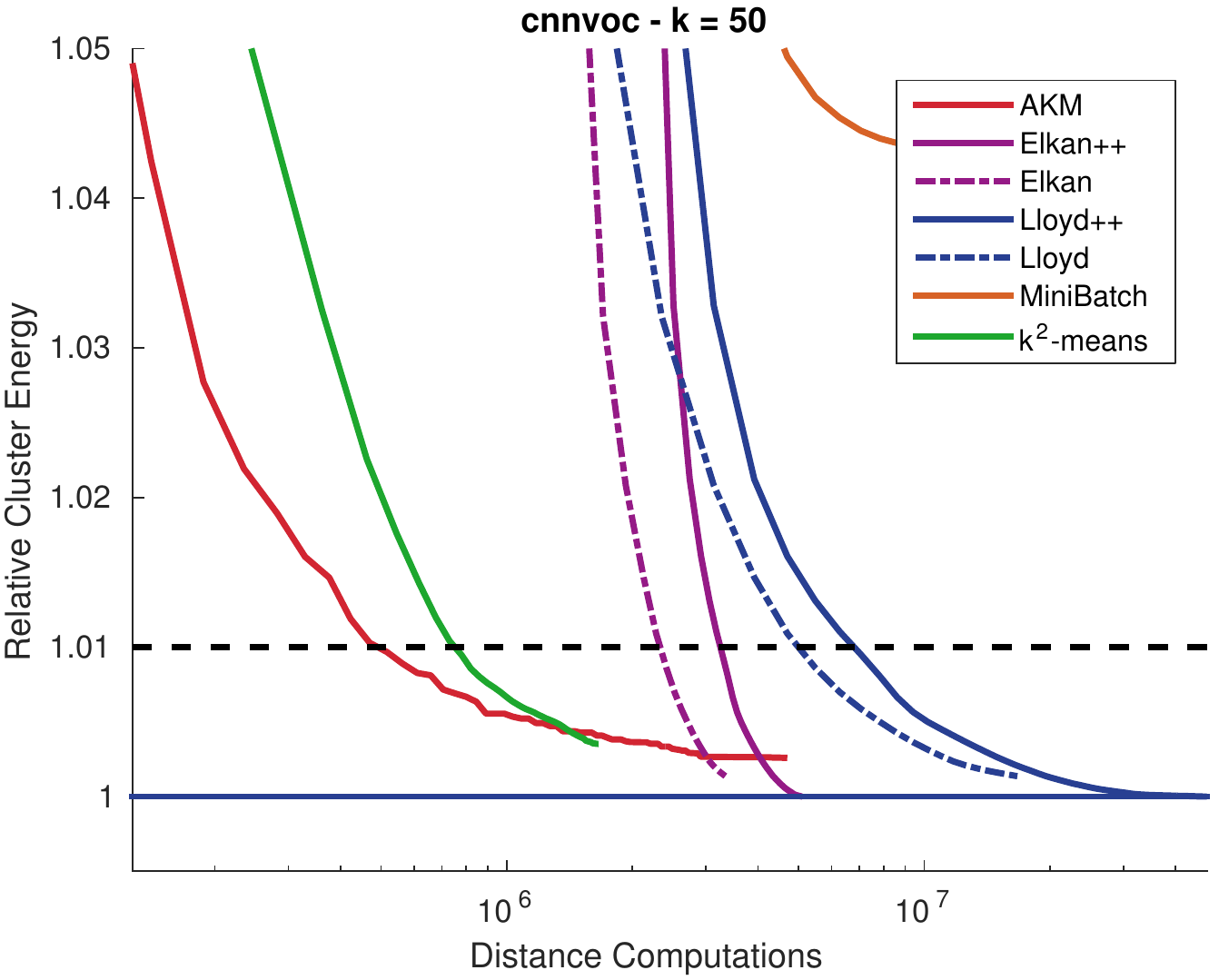} &
\includegraphics[width=0.32\textwidth]{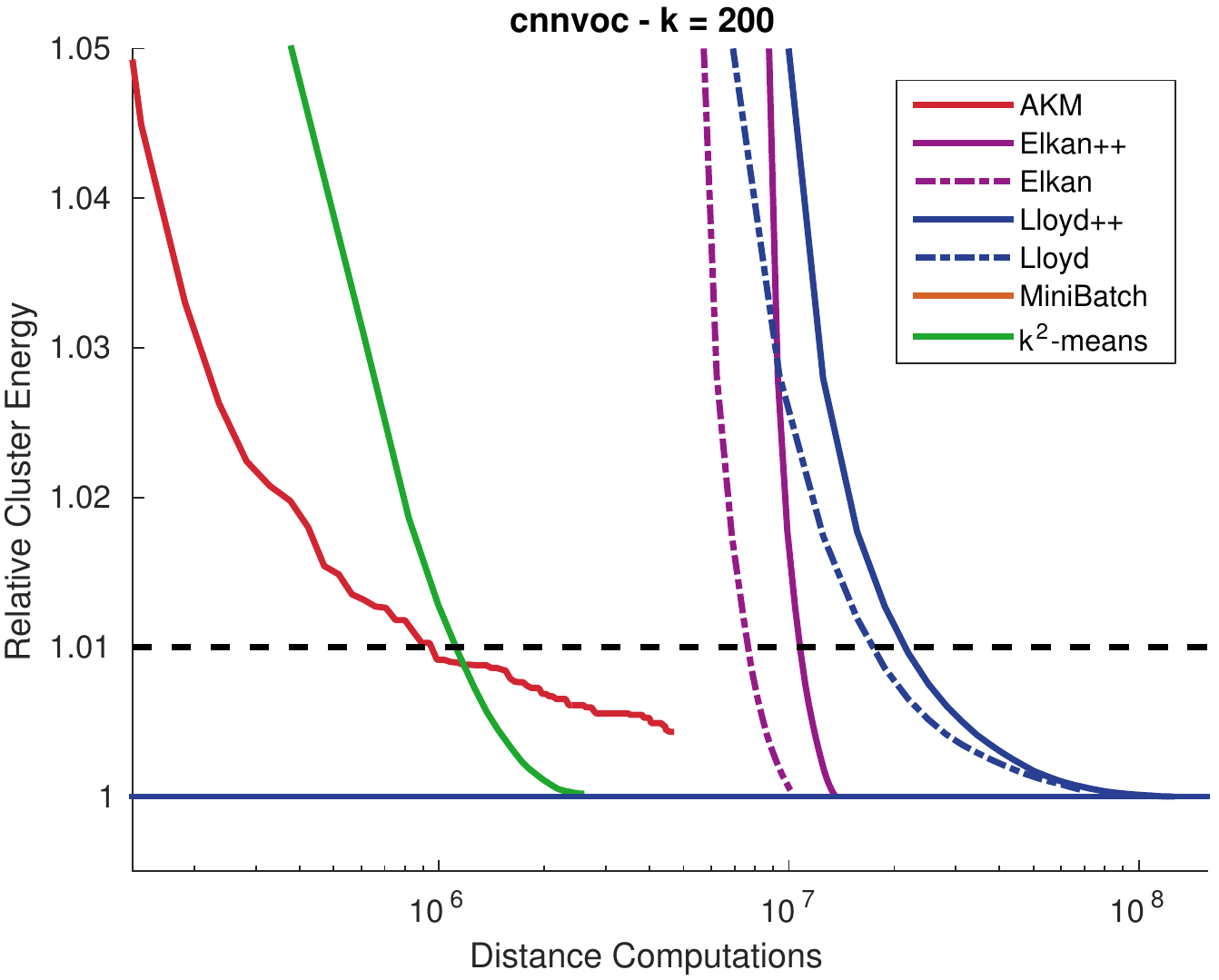} &
\includegraphics[width=0.32\textwidth]{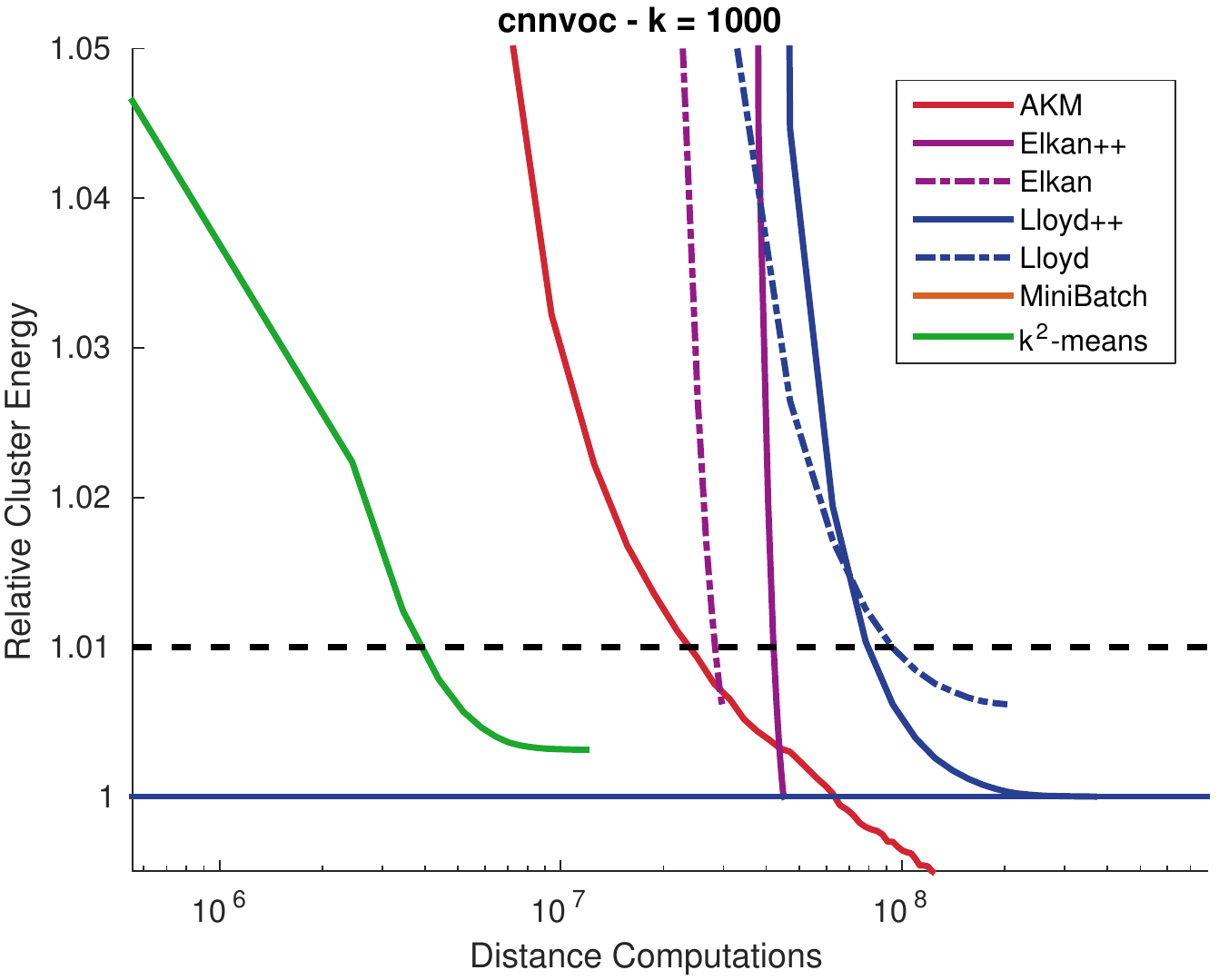} \\
\includegraphics[width=0.32\textwidth]{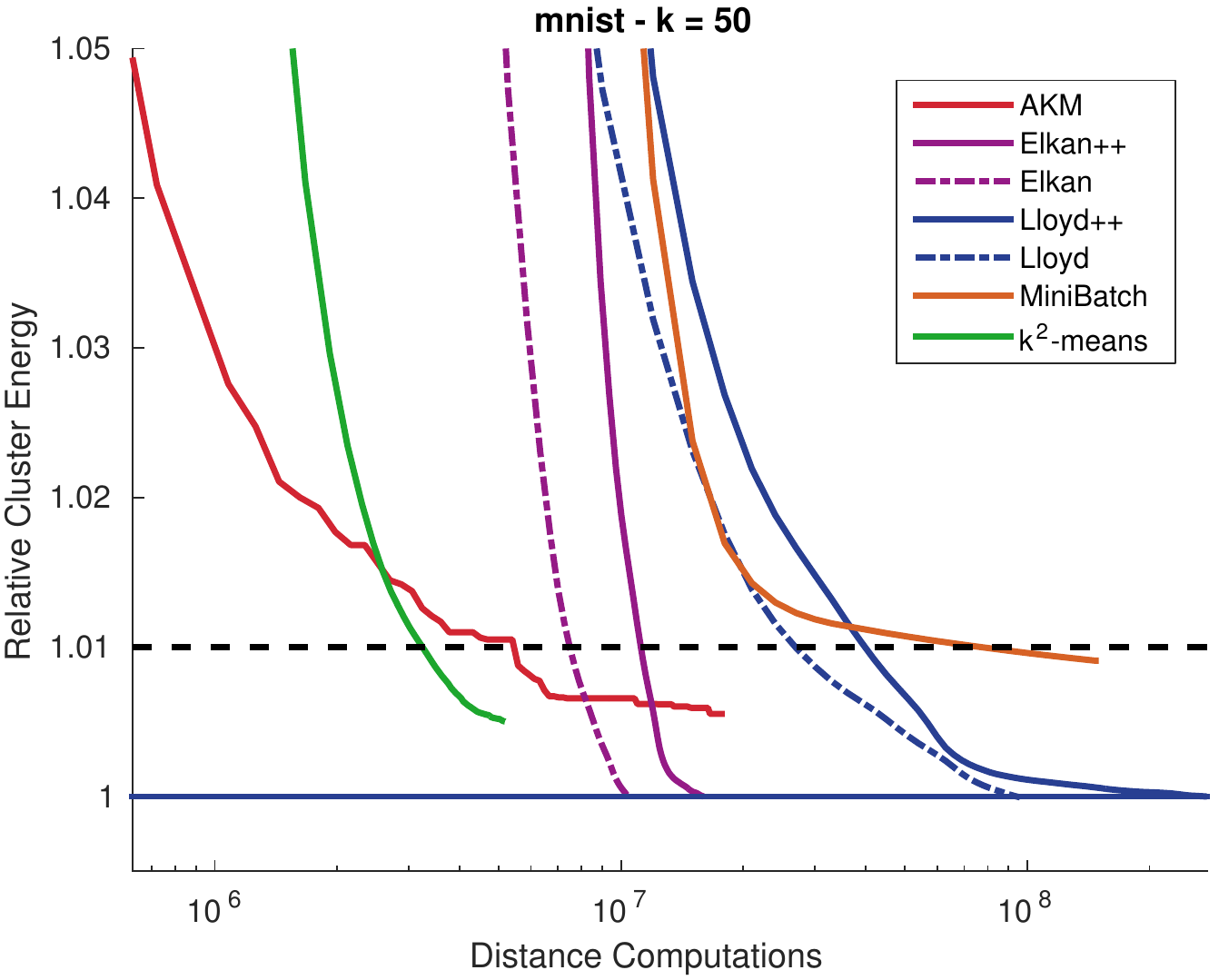} &
\includegraphics[width=0.32\textwidth]{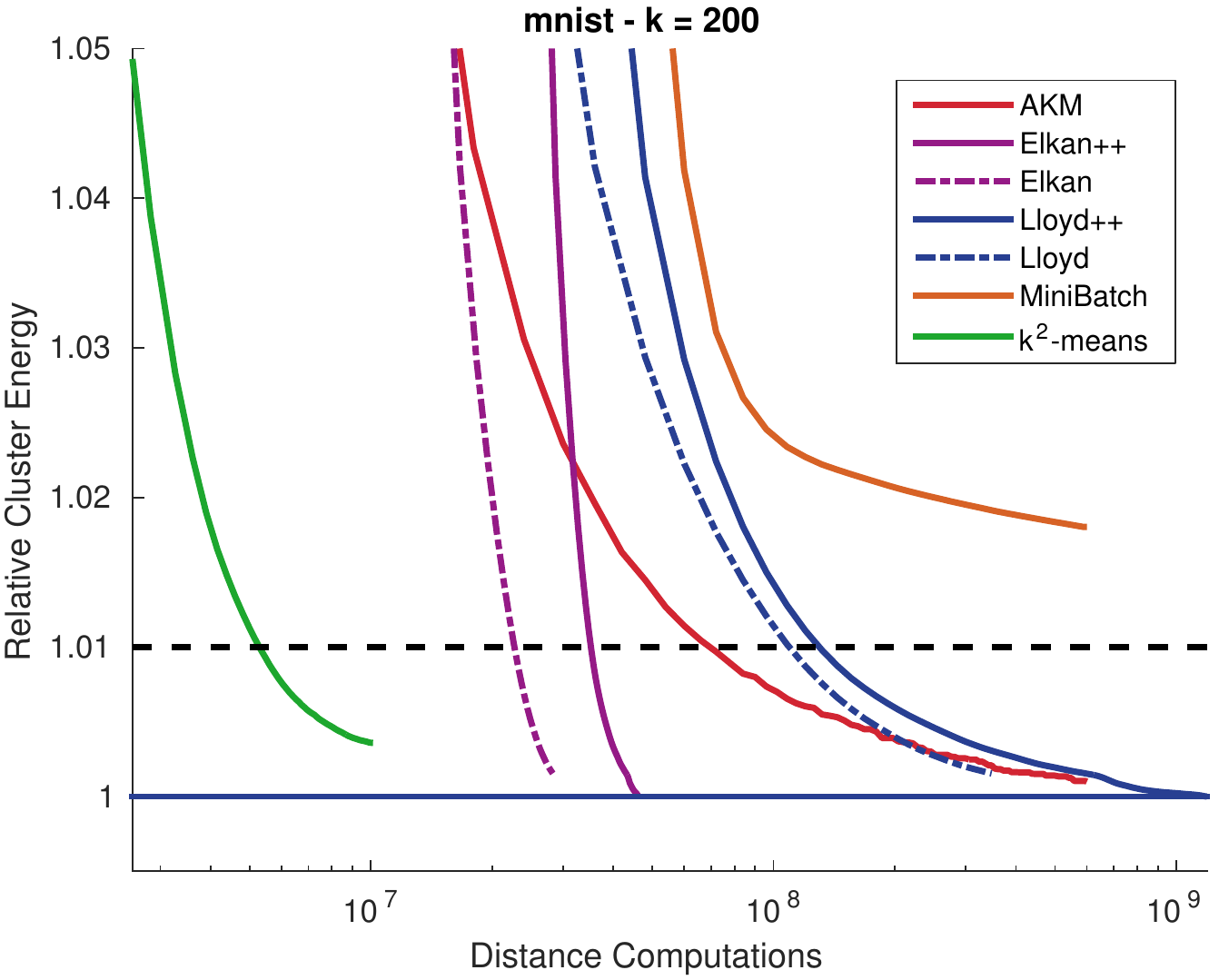} &
\includegraphics[width=0.32\textwidth]{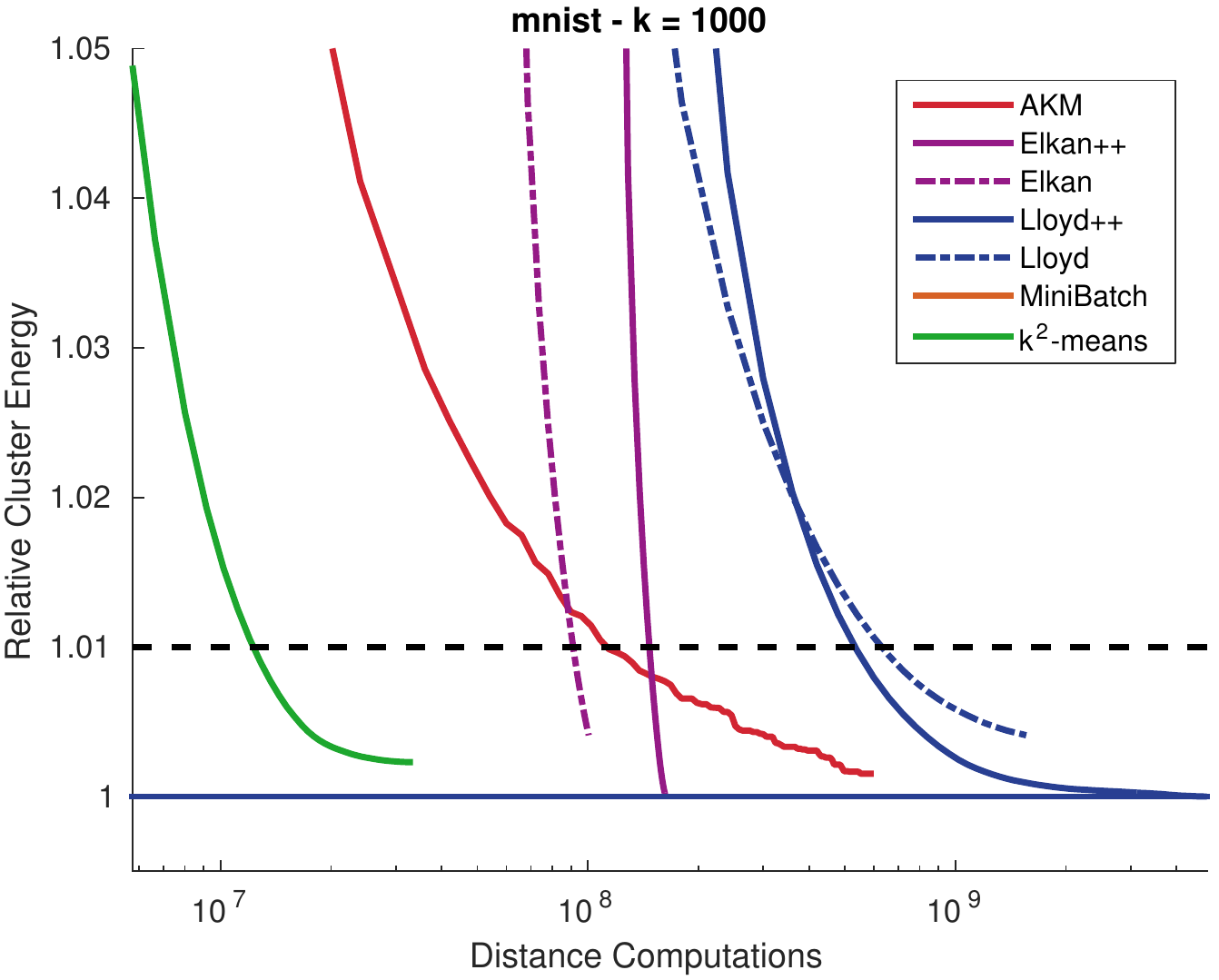} \\
\includegraphics[width=0.32\textwidth]{cluster_all_k2elkan_philbin_plusplus_mnist50_ce_dists_k=50.pdf} &
\includegraphics[width=0.32\textwidth]{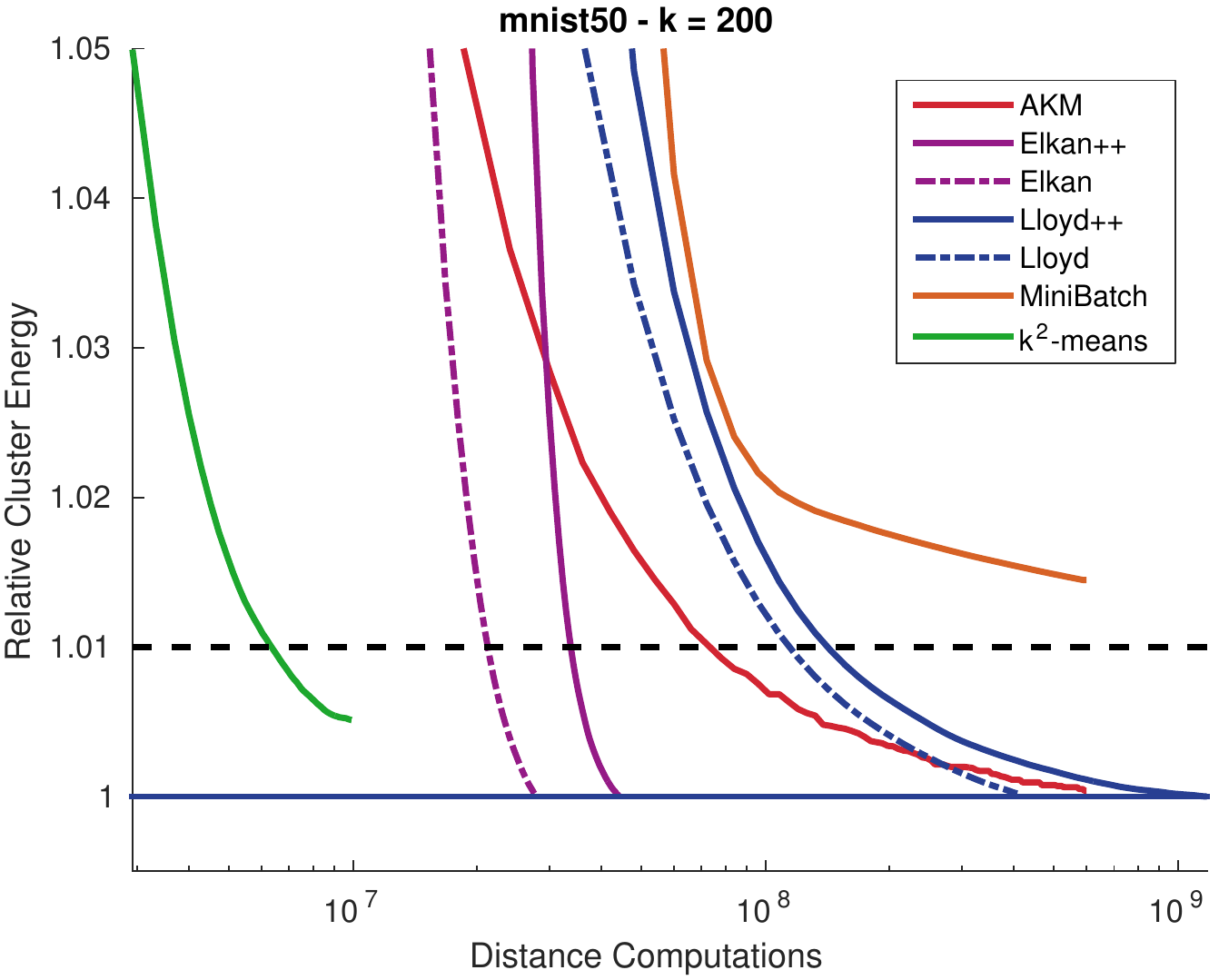} &
\includegraphics[width=0.32\textwidth]{cluster_all_k2elkan_philbin_plusplus_mnist50_ce_dists_k=1000.pdf} \\
\end{tabular}
}
\caption{Cluster Energy (relative to best Lloyd++ energy) vs distance computations on cifar, cnnvoc, mnist and mnist50 for $k\in\{50,200,1000\}$. For AKM and $k^2$-means, we use the parameter with the highest algorithmic speedup at 1\% error. }
\label{fig:k2means_plot}
\end{figure*}

\begin{figure*}
\begin{tabular}{ccc}
\includegraphics[width=0.32\textwidth]{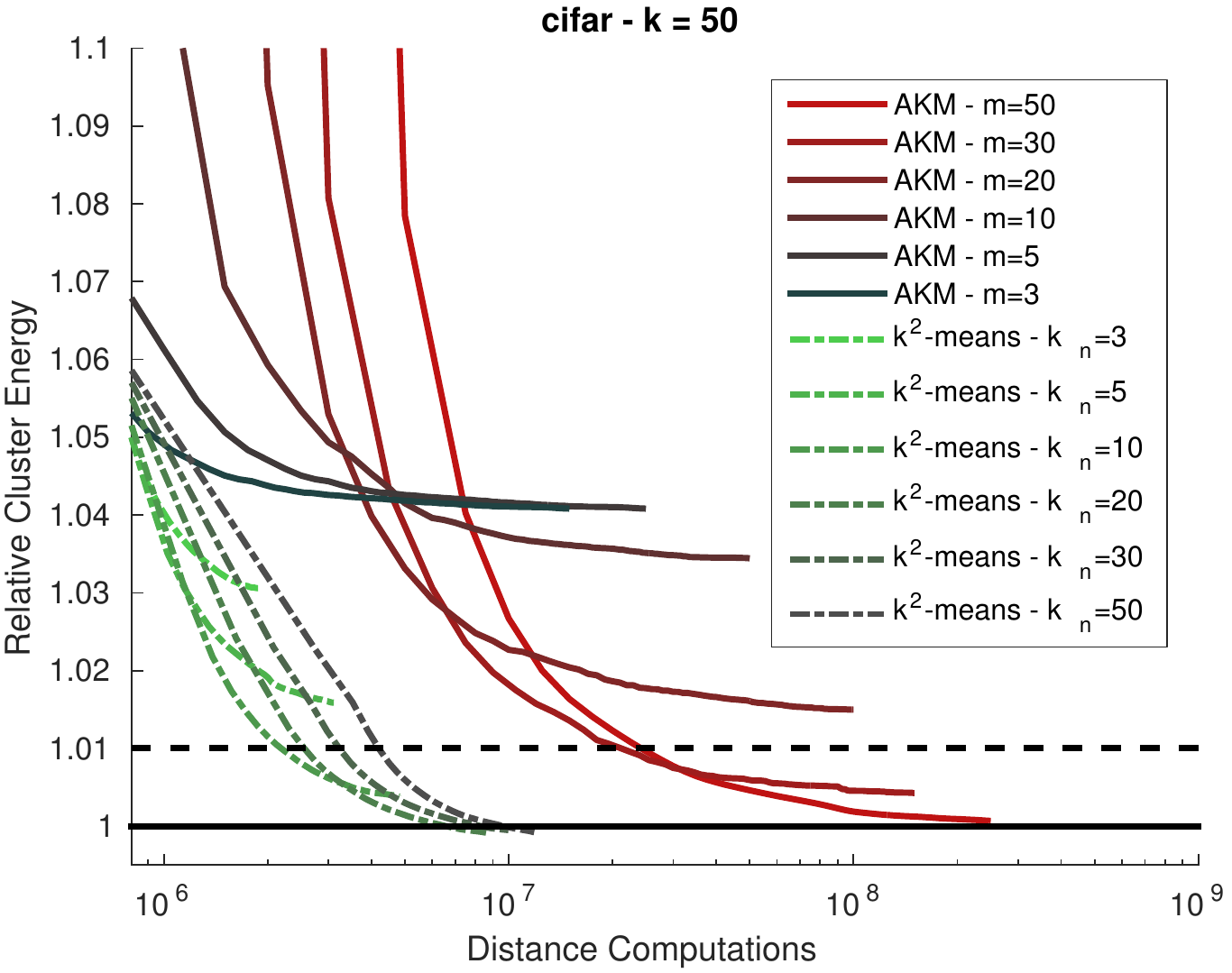} &
\includegraphics[width=0.32\textwidth]{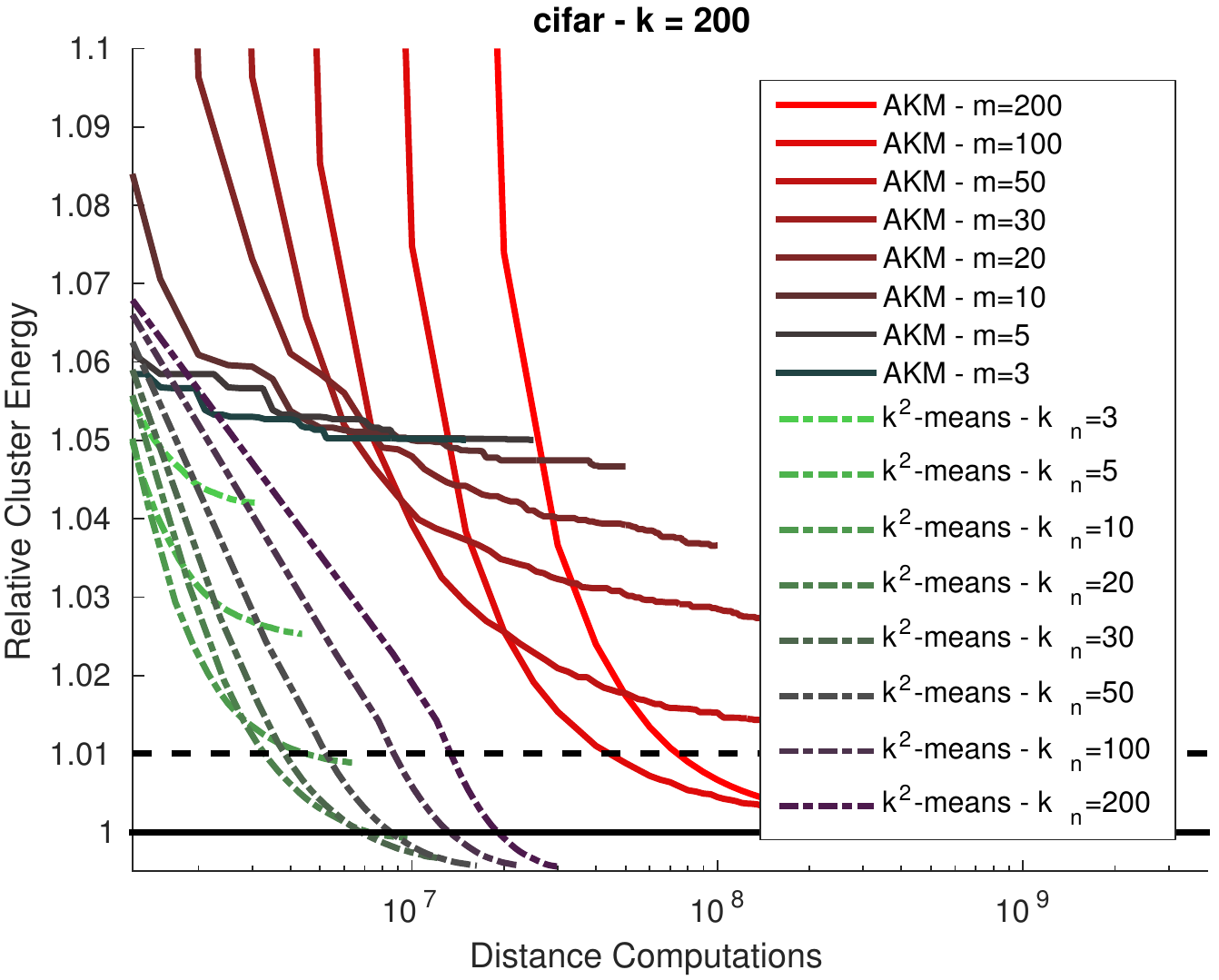} &
\includegraphics[width=0.32\textwidth]{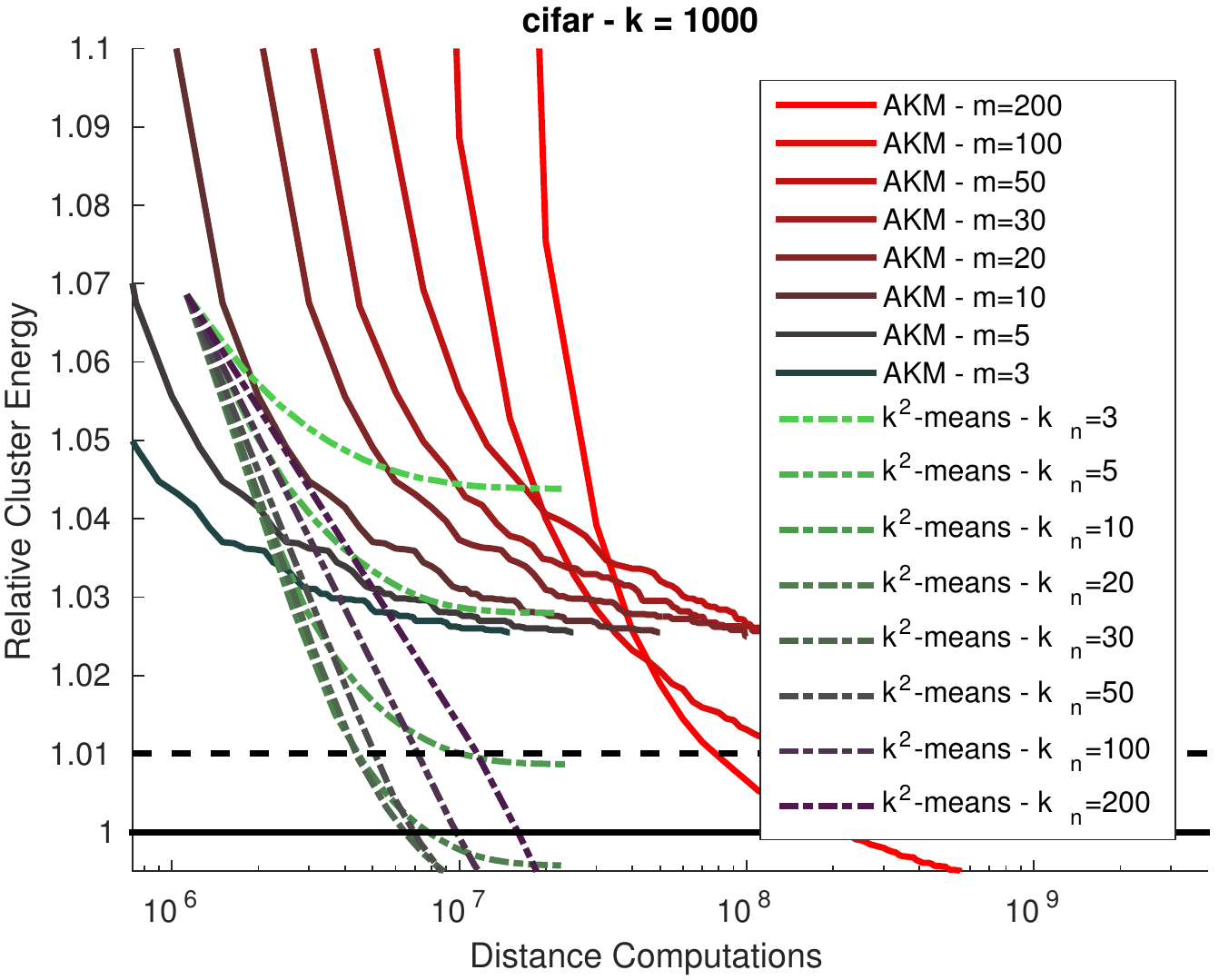} \\
\includegraphics[width=0.32\textwidth]{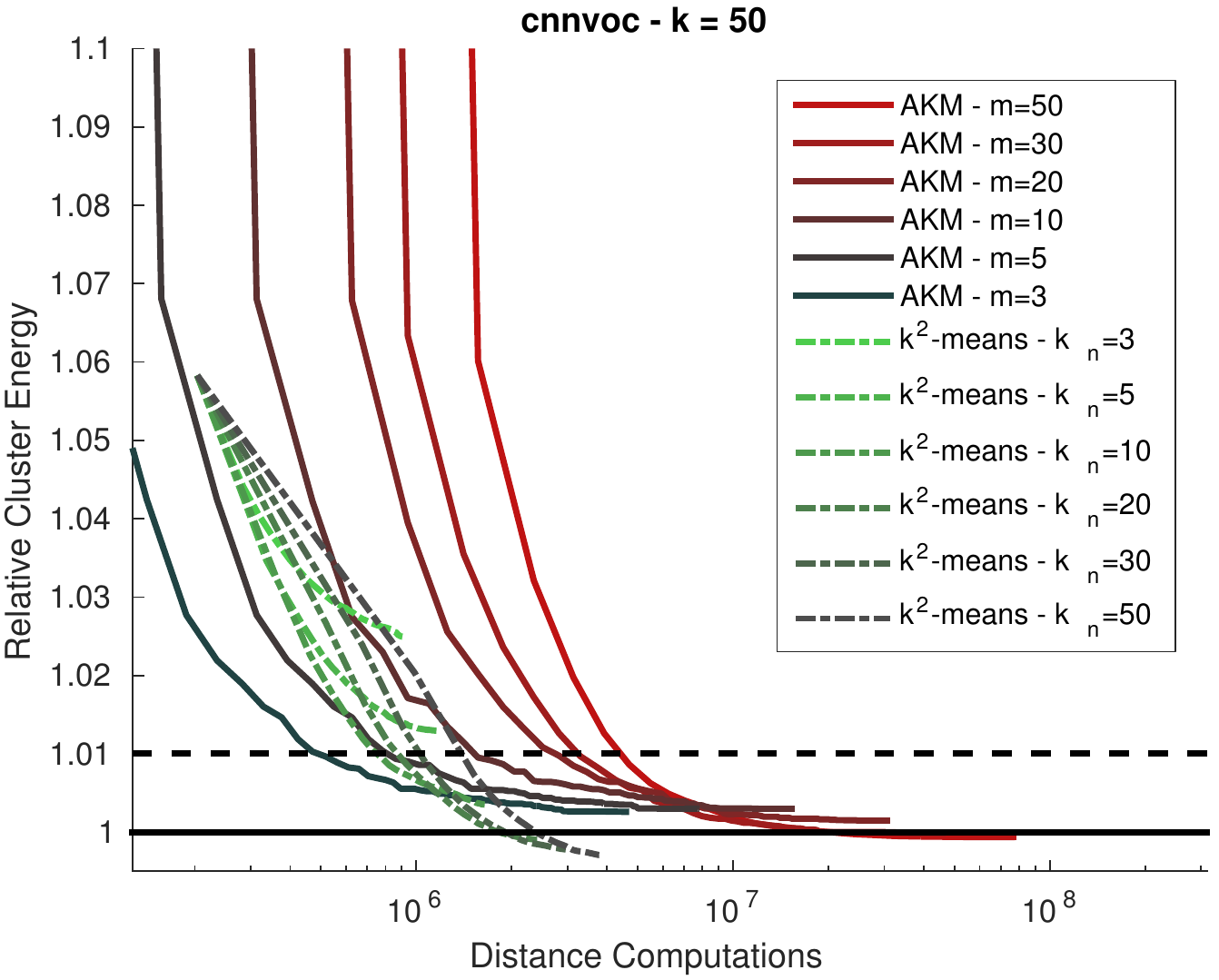} &
\includegraphics[width=0.32\textwidth]{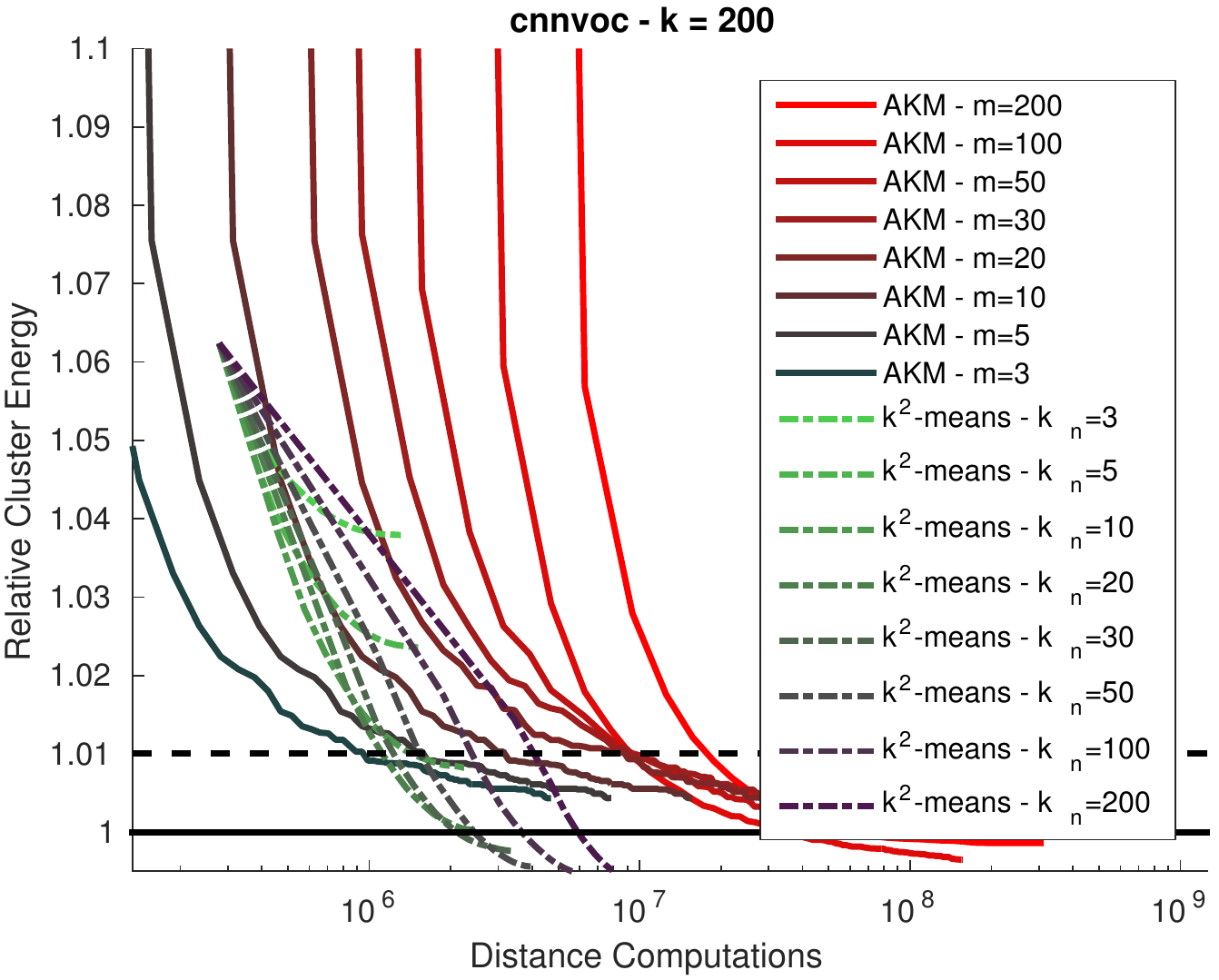} &
\includegraphics[width=0.32\textwidth]{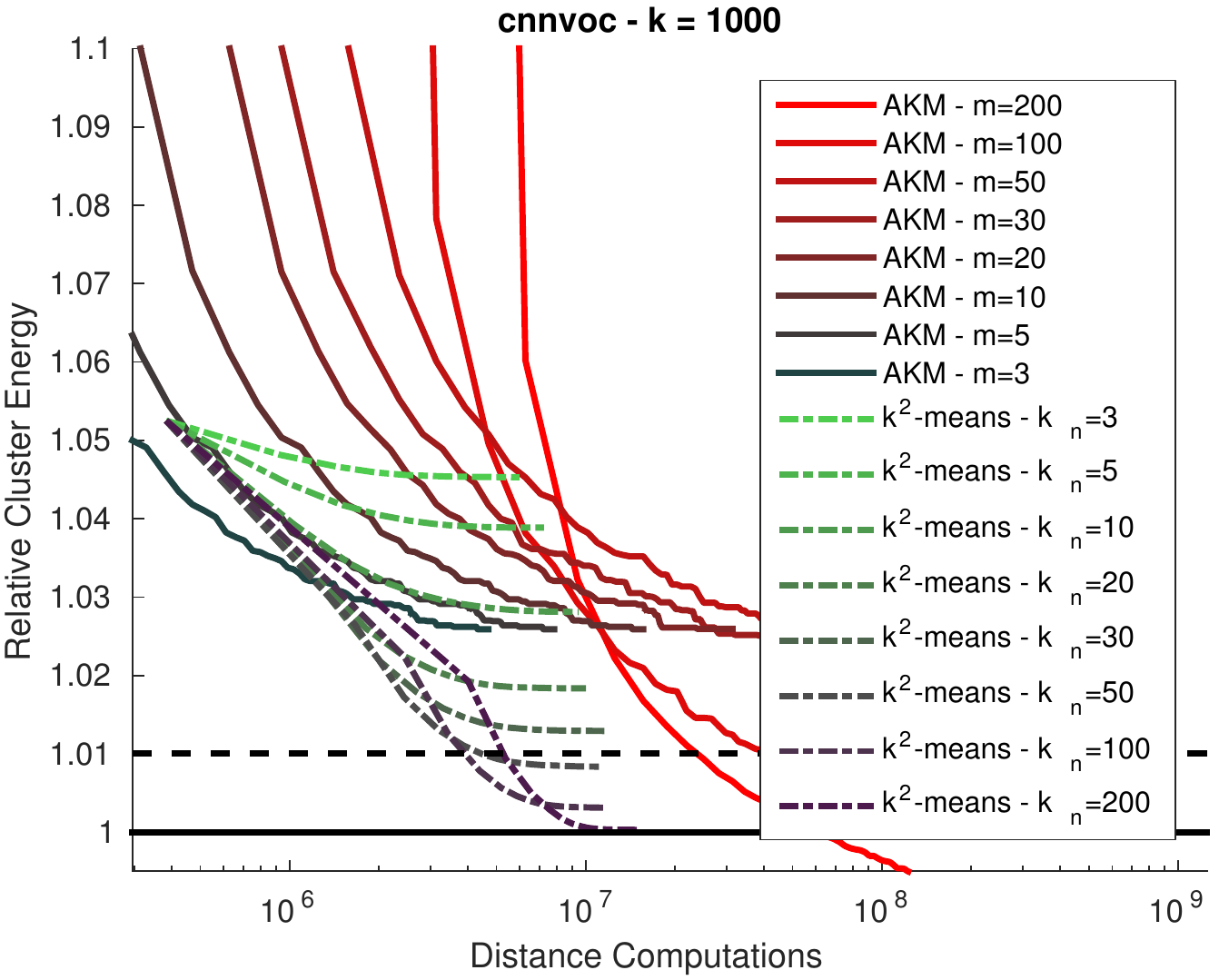} \\
\includegraphics[width=0.32\textwidth]{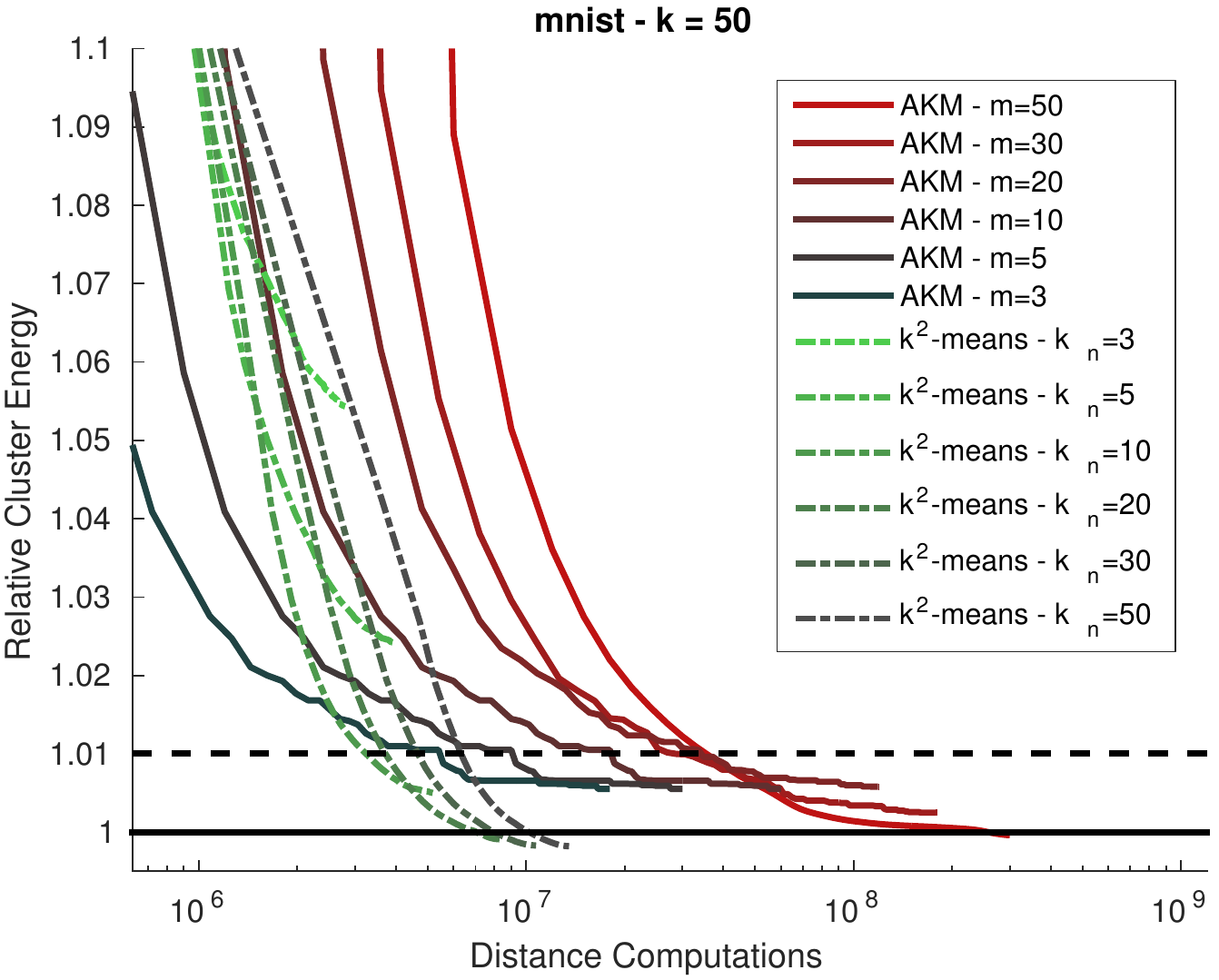} &
\includegraphics[width=0.32\textwidth]{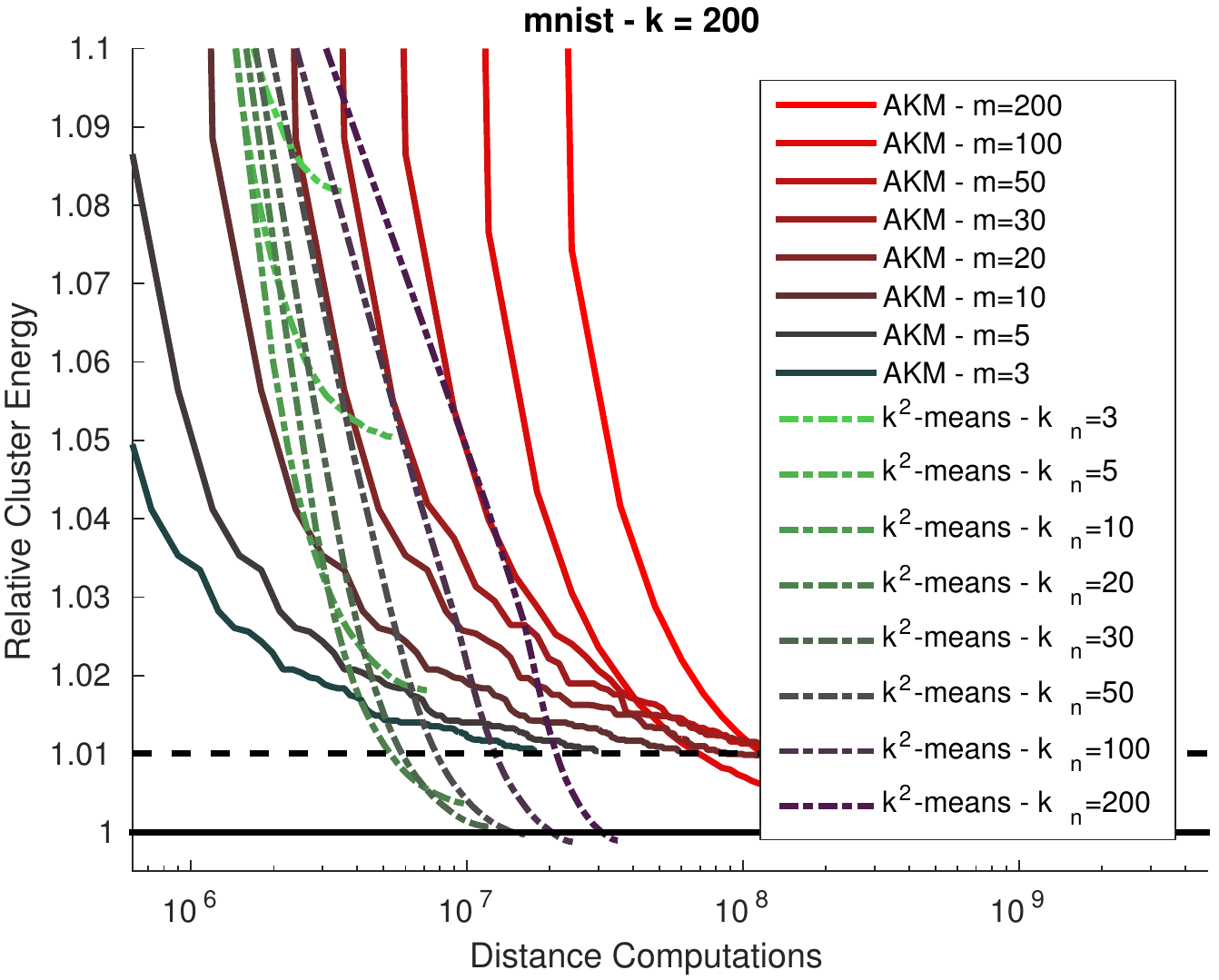} &
\includegraphics[width=0.32\textwidth]{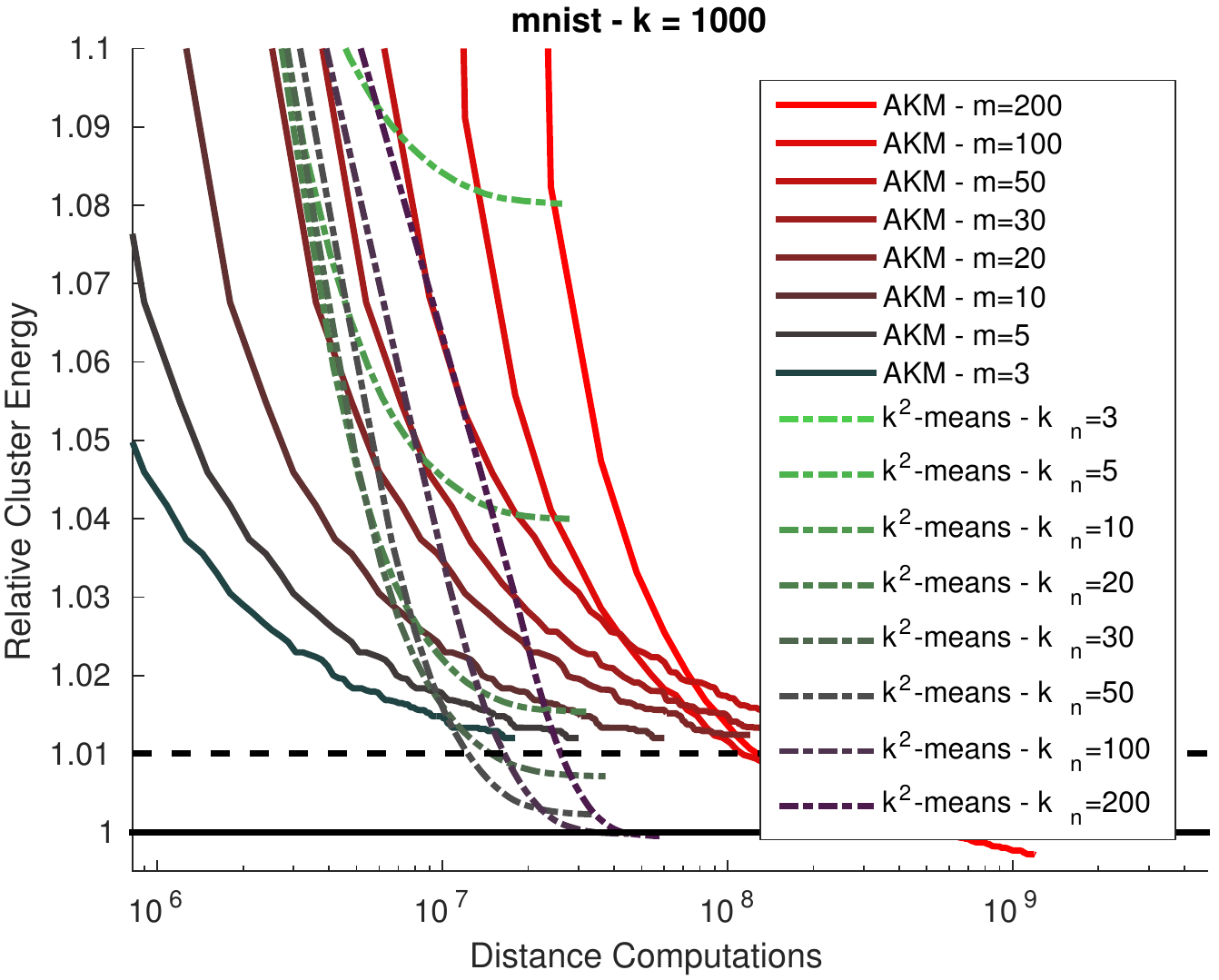} \\
\includegraphics[width=0.32\textwidth]{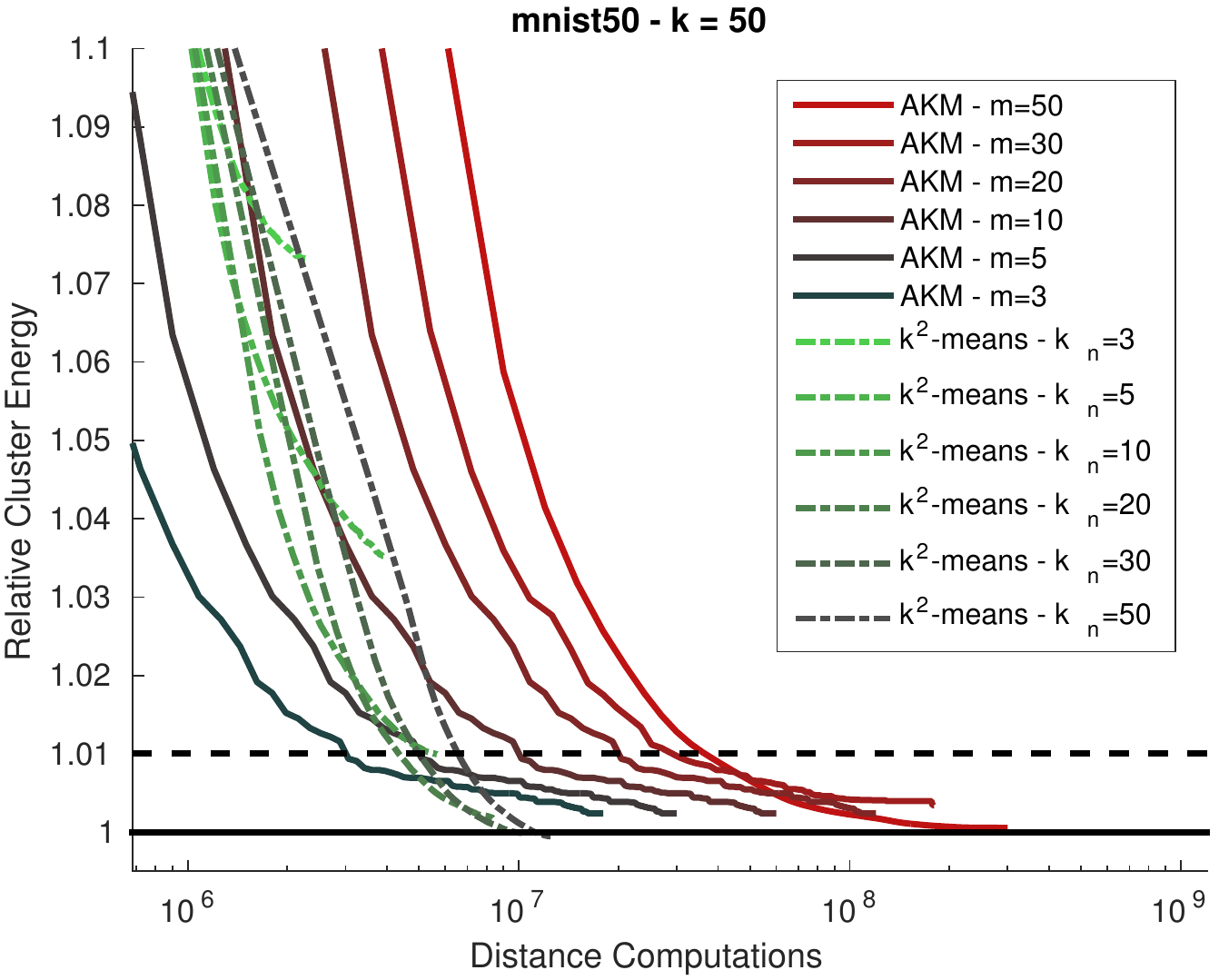} &
\includegraphics[width=0.32\textwidth]{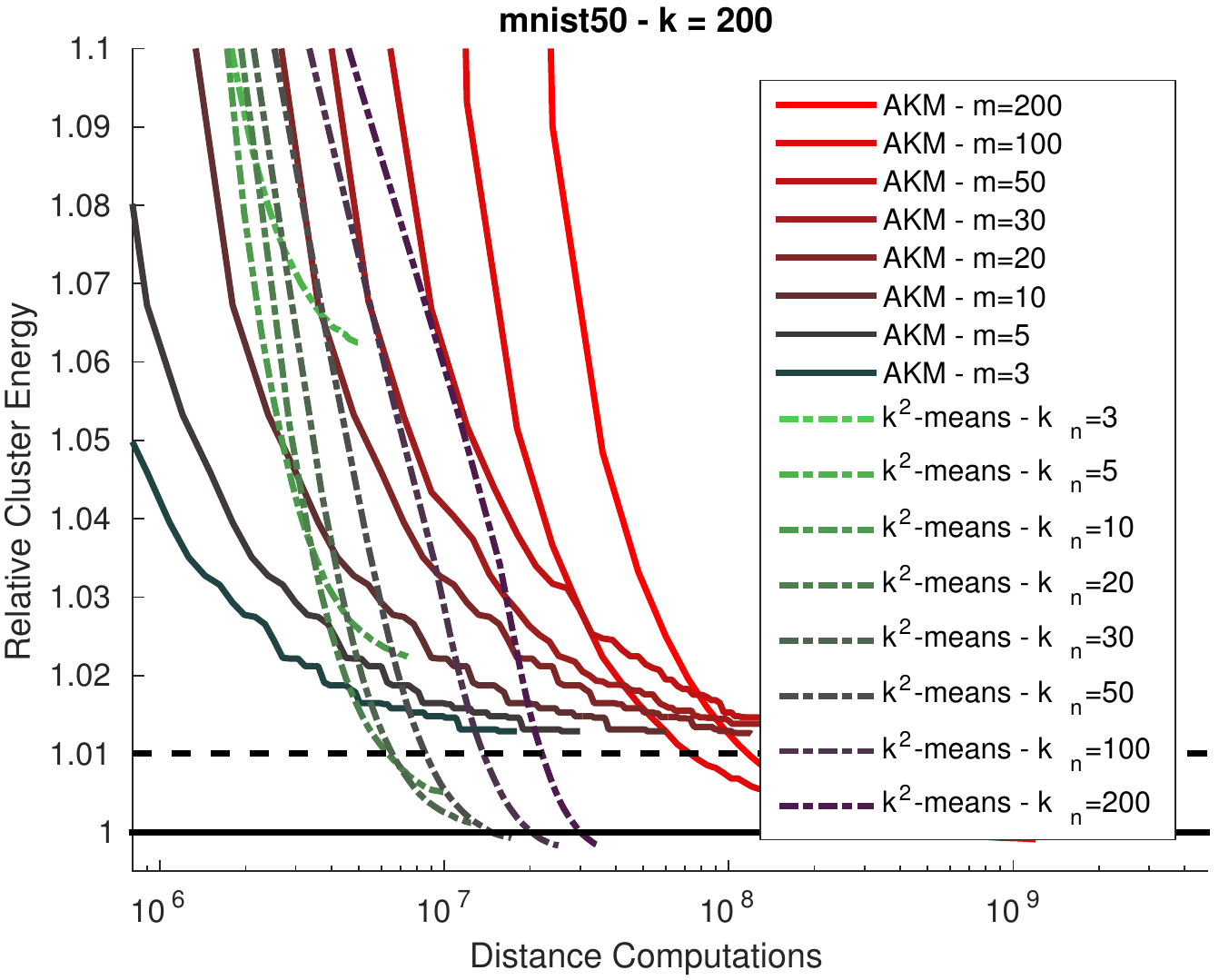} &
\includegraphics[width=0.32\textwidth]{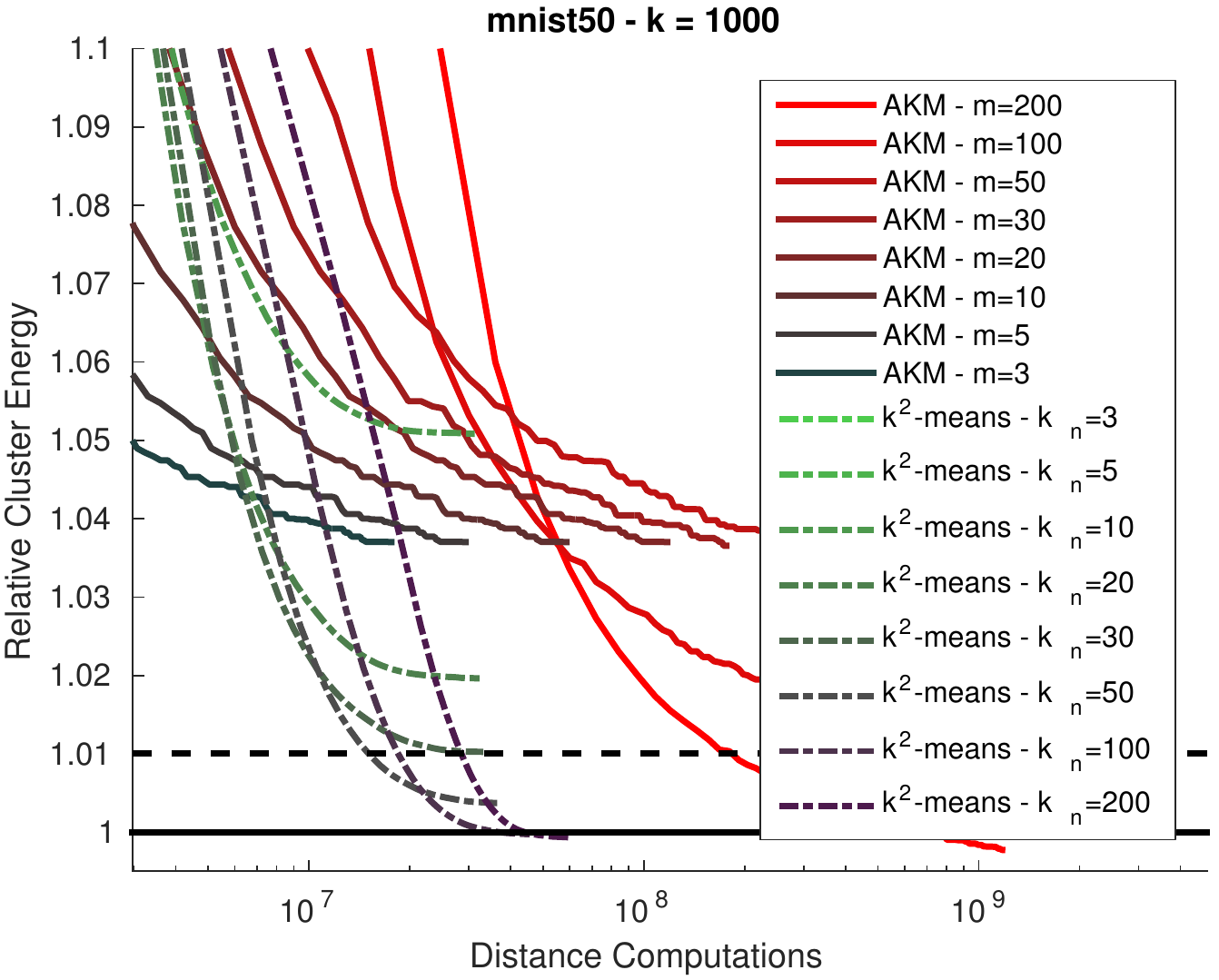} \\

\end{tabular}
\caption{Cluster Energy (relative to best Lloyd++ energy) vs distances computed for AKM and our $k^2$-means methods with different parameters and number of clusters, on the cifar, cnnvoc, mnist and mnist50 datasets.}
\label{fig:k2means_vs_ann_plot}
\end{figure*}

\begin{table*}
\begin{center}
\small
\tabcolsep=0.05cm
\begin{tabular}{|l|c|c|c|ccccccc|}
\hline
Dataset & $n$ & $d$ & $k$
& AKM & Elkan++ & Elkan & Lloyd++ & Lloyd & MiniBatch & $k^2$-means
\\
\hline\hline
cifar & 50000 & 3072 & 50 & - & 17.8 & - & 1.0 & - & - & \textbf{37.9} \\
& & & 100 & - & 20.7 & - & 1.0 & - & - & \textbf{70.6} \\
& & & 200 & 1.2 & 24.2 & - & 1.0 & - & - & \textbf{139.8} \\
& & & 500 & 9.0 & 20.5 & 31.9 & 1.0 & 2.9 & - & \textbf{287.9} \\
& & & 1000 & 11.3 & 17.5 & 28.2 & 1.0 & 2.6 & - & \textbf{373.6} \\
cnnvoc & 15662 & 4096 & 50 & 2.4 & 9.3 & - & 1.0 & - & - & \textbf{26.2} \\
& & & 100 & 1.3 & 11.0 & - & 1.0 & - & - & \textbf{44.3} \\
& & & 200 & 3.7 & 9.3 & - & 1.0 & - & - & \textbf{59.7} \\
& & & 500 & 5.7 & 9.2 & - & 1.0 & - & - & \textbf{79.2} \\
& & & 1000 & 5.8 & \textbf{8.1} & - & 1.0 & - & - & - \\
covtype & 150000 & 54 & 50 & - & 28.9 & - & 1.0 & - & - & \textbf{172.0} \\
& & & 100 & - & 34.3 & - & 1.0 & - & - & \textbf{287.2} \\
& & & 200 & - & 40.2 & - & 1.0 & - & - & \textbf{442.4} \\
& & & 500 & - & 47.4 & - & 1.0 & - & - & \textbf{816.9} \\
& & & 1000 & - & \textbf{44.5} & - & 1.0 & - & - & - \\
mnist & 60000 & 784 & 50 & 1.1 & 17.3 & 26.6 & 1.0 & 2.9 & - & \textbf{39.3} \\
& & & 100 & 2.7 & 20.5 & - & 1.0 & - & - & \textbf{56.8} \\
& & & 200 & - & 25.8 & - & 1.0 & - & - & \textbf{81.0} \\
& & & 500 & 5.0 & 23.6 & - & 1.0 & - & - & \textbf{108.1} \\
& & & 1000 & 9.0 & 29.8 & - & 1.0 & - & - & \textbf{141.1} \\
mnist50 & 60000 & 50 & 50 & - & 18.7 & - & 1.0 & - & - & \textbf{31.0} \\
& & & 100 & 1.4 & 22.3 & - & 1.0 & - & - & \textbf{39.7} \\
& & & 200 & 2.1 & 26.6 & - & 1.0 & - & - & \textbf{80.3} \\
& & & 500 & 3.4 & 23.2 & - & 1.0 & - & - & \textbf{79.8} \\
& & & 1000 & 5.1 & 22.7 & - & 1.0 & - & - & \textbf{94.1} \\
tinygist10k & 10000 & 384 & 50 & 12.5 & 12.5 & 20.1 & 1.0 & 3.6 & - & \textbf{50.1} \\
& & & 100 & 3.0 & 11.7 & 17.0 & 1.0 & 2.8 & - & \textbf{53.5} \\
& & & 200 & 4.7 & 11.0 & - & 1.0 & - & - & \textbf{71.8} \\
& & & 500 & 6.2 & 10.7 & - & 1.0 & - & - & \textbf{71.7} \\
& & & 1000 & 2.6 & \textbf{7.8} & - & 1.0 & - & - & - \\
usps & 7291 & 256 & 50 & - & 12.6 & - & 1.0 & - & - & \textbf{31.7} \\
& & & 100 & - & 15.0 & - & 1.0 & - & - & \textbf{46.4} \\
& & & 200 & 3.4 & 14.6 & - & 1.0 & - & - & \textbf{54.4} \\
& & & 500 & - & 10.8 & - & 1.0 & - & - & \textbf{47.6} \\
& & & 1000 & - & \textbf{9.4} & - & 1.0 & - & - & - \\
yale & 2414 & 32256 & 50 & 2.8 & 9.5 & - & 1.0 & - & - & \textbf{32.5} \\
& & & 100 & - & \textbf{8.0} & - & 1.0 & - & - & - \\
& & & 200 & \textbf{20.8} & 6.5 & - & 1.0 & - & - & 18.7 \\
& & & 500 & \textbf{21.6} & 5.3 & - & 1.0 & - & - & - \\
& & & 1000 & - & \textbf{4.0} & - & 1.0 & - & - & - \\

\hline
\end{tabular}
\end{center}
\caption{Algorithmic speedup in reaching the same energy as the final Lloyd++ energy. (-) marks failure in reaching the target of $0\%$ relative error.
For each method, the parameter(s) that gave the highest speedup at 0\% error is used.  }
\label{tab:k2means_speedup00_exp}
\end{table*}

\begin{table*}
\begin{center}
\small
\tabcolsep=0.05cm
\begin{tabular}{|l|c|c|c|ccccccc|}
\hline
Dataset & $n$ & $d$ & $k$
& AKM & Elkan++ & Elkan & Lloyd++ & Lloyd &MiniBatch  & $k^2$-means
\\
\hline\hline
cifar & 50000 & 3072 & 50 & 0.8 & 4.1 & 5.6 & 1.0 & 1.1 & - & \textbf{10.8} \\
& & & 100 & 0.7 & 4.7 & 7.2 & 1.0 & 1.4 & - & \textbf{20.1} \\
& & & 200 & 1.6 & 4.5 & 6.7 & 1.0 & 1.2 & - & \textbf{32.4} \\
& & & 500 & 3.0 & 4.1 & 6.5 & 1.0 & 1.2 & - & \textbf{69.0} \\
& & & 1000 & 4.6 & 4.1 & 6.7 & 1.0 & 1.2 & - & \textbf{101.0} \\
cnnvoc & 15662 & 4096 & 50 & \textbf{8.8} & 2.8 & 3.8 & 1.0 & 1.3 & - & 8.7 \\
& & & 100 & 4.0 & 2.6 & 3.3 & 1.0 & 1.0 & - & \textbf{12.9} \\
& & & 200 & 7.8 & 2.7 & 3.7 & 1.0 & 1.2 & - & \textbf{21.9} \\
& & & 500 & 2.7 & 2.4 & 3.4 & 1.0 & 1.0 & - & \textbf{26.0} \\
& & & 1000 & 2.9 & 2.4 & - & 1.0 & - & - & \textbf{18.2} \\
covtype & 150000 & 54 & 50 & - & 9.8 & - & 1.0 & - & - & \textbf{56.8} \\
& & & 100 & - & 7.5 & - & 1.0 & - & - & \textbf{62.4} \\
& & & 200 & - & 8.1 & - & 1.0 & - & - & \textbf{96.8} \\
& & & 500 & - & 12.5 & - & 1.0 & - & - & \textbf{249.8} \\
& & & 1000 & - & 12.0 & - & 1.0 & - & - & \textbf{190.0} \\
mnist & 60000 & 784 & 50 & 1.0 & 4.7 & 6.6 & 1.0 & 1.2 & - & \textbf{12.4} \\
& & & 100 & 1.2 & 4.9 & 7.1 & 1.0 & 1.2 & - & \textbf{18.3} \\
& & & 200 & 1.5 & 6.1 & 9.3 & 1.0 & 1.3 & - & \textbf{31.8} \\
& & & 500 & 2.9 & 5.5 & 8.6 & 1.0 & 1.1 & - & \textbf{44.5} \\
& & & 1000 & 3.7 & 4.9 & 7.7 & 1.0 & 0.6 & - & \textbf{46.3} \\
mnist50 & 60000 & 50 & 50 & 7.5 & 5.4 & 7.6 & 1.0 & 1.2 & - & \textbf{11.1} \\
& & & 100 & 1.1 & 6.1 & 9.0 & 1.0 & 1.2 & - & \textbf{18.7} \\
& & & 200 & 1.8 & 6.7 & 10.5 & 1.0 & 1.4 & - & \textbf{30.1} \\
& & & 500 & 1.9 & 5.8 & 9.0 & 1.0 & 0.8 & - & \textbf{35.6} \\
& & & 1000 & 2.6 & 5.5 & 8.7 & 1.0 & 0.6 & - & \textbf{36.7} \\
tinygist10k & 10000 & 384 & 50 & 11.6 & 3.4 & 5.1 & 1.0 & 1.6 & - & \textbf{14.2} \\
& & & 100 & 7.9 & 3.1 & 4.4 & 1.0 & 1.2 & - & \textbf{16.3} \\
& & & 200 & 7.8 & 3.1 & 4.6 & 1.0 & 1.3 & - & \textbf{24.8} \\
& & & 500 & 2.3 & 2.6 & 4.0 & 1.0 & 0.8 & - & \textbf{25.3} \\
& & & 1000 & 1.3 & \textbf{2.5} & - & 1.0 & - & - & - \\
usps & 7291 & 256 & 50 & - & 5.8 & - & 1.0 & - & - & \textbf{15.4} \\
& & & 100 & 0.7 & 6.5 & - & 1.0 & - & - & \textbf{23.5} \\
& & & 200 & 17.0 & 5.8 & - & 1.0 & - & - & \textbf{25.2} \\
& & & 500 & 15.6 & 4.1 & - & 1.0 & - & - & \textbf{23.5} \\
& & & 1000 & - & \textbf{3.3} & - & 1.0 & - & - & - \\
yale & 2414 & 32256 & 50 & 2.6 & 6.0 & - & 1.0 & - & - & \textbf{23.6} \\
& & & 100 & 1.1 & 3.6 & - & 1.0 & - & - & \textbf{12.7} \\
& & & 200 & \textbf{14.2} & 3.5 & - & 1.0 & - & - & 13.7 \\
& & & 500 & \textbf{14.4} & 2.6 & - & 1.0 & - & - & - \\
& & & 1000 & - & \textbf{2.0} & - & 1.0 & - & - & - \\

\hline
\end{tabular}
\end{center}
\caption{Algorithmic speedup in reaching an energy within $0.5\%$ from the final Lloyd++ energy. (-) marks failure in reaching the target of $0.5\%$ relative error.
For each method, the parameter(s) that gave the highest speedup at 0.5\% error is used. }
\label{tab:k2means_speedup5_exp}
\end{table*}

\begin{table*}
\begin{center}
\small
\tabcolsep=0.05cm
\begin{tabular}{|l|c|c|c|ccccccc|}
\hline
Dataset & $n$ & $d$ & $k$
& AKM & Elkan++ & Elkan & Lloyd++ & Lloyd &MiniBatch  & $k^2$-means
\\
\hline\hline
cifar & 50000 & 3072 & 50 & 1.0 & 2.6 & 3.7 & 1.0 & 1.0 & - & \textbf{9.5} \\
& & & 100 & 1.0 & 3.1 & 4.8 & 1.0 & 1.3 & - & \textbf{15.7} \\
& & & 200 & 1.9 & 3.0 & 4.6 & 1.0 & 1.1 & - & \textbf{26.2} \\
& & & 500 & 3.1 & 3.0 & 4.9 & 1.0 & 1.2 & - & \textbf{59.0} \\
& & & 1000 & 4.9 & 3.0 & 5.1 & 1.0 & 1.2 & - & \textbf{86.7} \\
cnnvoc & 15662 & 4096 & 50 & \textbf{13.8} & 2.1 & 2.9 & 1.0 & 1.4 & - & 9.0 \\
& & & 100 & \textbf{14.2} & 2.0 & 2.6 & 1.0 & 1.1 & - & 11.2 \\
& & & 200 & \textbf{22.6} & 2.0 & 2.8 & 1.0 & 1.2 & - & 19.2 \\
& & & 500 & 15.1 & 1.9 & 2.8 & 1.0 & 1.1 & - & \textbf{25.8} \\
& & & 1000 & 3.3 & 1.9 & 2.8 & 1.0 & 0.9 & - & \textbf{20.2} \\
covtype & 150000 & 54 & 50 & - & 6.1 & - & 1.0 & - & - & \textbf{35.1} \\
& & & 100 & - & 5.6 & - & 1.0 & - & - & \textbf{46.9} \\
& & & 200 & - & 6.3 & - & 1.0 & - & - & \textbf{78.7} \\
& & & 500 & - & 8.1 & - & 1.0 & - & - & \textbf{175.8} \\
& & & 1000 & - & 8.5 & - & 1.0 & - & - & \textbf{176.6} \\
mnist & 60000 & 784 & 50 & 7.3 & 3.6 & 5.3 & 1.0 & 1.5 & 0.5 & \textbf{12.3} \\
& & & 100 & 4.2 & 3.3 & 4.9 & 1.0 & 1.1 & - & \textbf{14.6} \\
& & & 200 & 1.9 & 3.7 & 5.7 & 1.0 & 1.2 & - & \textbf{24.6} \\
& & & 500 & 3.5 & 3.8 & 6.2 & 1.0 & 1.1 & - & \textbf{39.8} \\
& & & 1000 & 4.7 & 3.6 & 5.9 & 1.0 & 0.8 & - & \textbf{43.4} \\
mnist50 & 60000 & 50 & 50 & \textbf{12.7} & 3.7 & 5.4 & 1.0 & 1.3 & - & 8.8 \\
& & & 100 & 8.3 & 4.1 & 6.2 & 1.0 & 1.2 & - & \textbf{15.0} \\
& & & 200 & 1.9 & 4.2 & 6.7 & 1.0 & 1.2 & - & \textbf{22.3} \\
& & & 500 & 2.2 & 4.2 & 6.8 & 1.0 & 1.0 & - & \textbf{33.1} \\
& & & 1000 & 3.1 & 4.1 & 6.6 & 1.0 & 0.8 & - & \textbf{38.0} \\
tiny10k & 10000 & 3072 & 50 & 1.5 & 2.9 & 4.0 & 1.0 & 1.0 & - & \textbf{12.3} \\
& & & 100 & 1.3 & 2.7 & 4.1 & 1.0 & 1.2 & - & \textbf{21.0} \\
& & & 200 & 2.0 & 2.9 & 4.5 & 1.0 & 1.1 & - & \textbf{35.4} \\
& & & 500 & 2.3 & 2.7 & 4.4 & 1.0 & 0.6 & - & \textbf{49.0} \\
& & & 1000 & 2.2 & 2.6 & - & 1.0 & - & - & \textbf{51.8} \\
tinygist10k & 10000 & 384 & 50 & \textbf{16.2} & 2.4 & 3.6 & 1.0 & 1.4 & - & 11.7 \\
& & & 100 & 12.9 & 2.2 & 3.3 & 1.0 & 1.2 & - & \textbf{14.2} \\
& & & 200 & 12.8 & 2.3 & 3.5 & 1.0 & 1.3 & - & \textbf{22.3} \\
& & & 500 & 2.5 & 2.1 & 3.3 & 1.0 & 1.0 & - & \textbf{26.2} \\
& & & 1000 & 1.5 & 2.1 & - & 1.0 & - & - & \textbf{13.6} \\
usps & 7291 & 256 & 50 & 5.3 & 4.1 & - & 1.0 & - & - & \textbf{11.8} \\
& & & 100 & 7.2 & 4.7 & 7.1 & 1.0 & 0.6 & - & \textbf{19.0} \\
& & & 200 & 16.8 & 4.4 & - & 1.0 & - & - & \textbf{23.6} \\
& & & 500 & 19.6 & 3.3 & - & 1.0 & - & - & \textbf{23.8} \\
& & & 1000& \textbf{18.5} & 2.7 & - & 1.0 & - & - & - \\
yale & 2414& 32256 & 50 & 2.1 & 4.2 & 6.3 & 1.0 & 0.6 & - & \textbf{17.9} \\
& & & 100 & 1.2 & 2.9 & - & 1.0 & - & - & \textbf{12.0} \\
& & & 200 & \textbf{21.9} & 2.9 & - & 1.0 & - & - & 13.9 \\
& & & 500 & \textbf{18.1} & 2.3 & - & 1.0 & - & - & - \\
& & & 1000 & - & \textbf{1.9} & - & 1.0 & - & - & - \\

\hline
\end{tabular}
\end{center}
\caption{Algorithmic speedup in reaching an energy within $1\%$ from the final Lloyd++ energy. (-) marks failure in reaching the target of $1\%$ relative error.
For each method, the parameter(s) that gave the highest speedup at 1\% error is used. }
\label{tab:k2means_speedup10_exp}
\end{table*}

\begin{table*}
\begin{center}
\small
\tabcolsep=0.05cm
\begin{tabular}{|l|c|c|c|ccccccc|}
\hline
Dataset & $n$ & $d$ & $k$
& AKM & Elkan++ & Elkan & Lloyd++ & Lloyd & MiniBatch & $k^2$-means
\\
\hline\hline
cifar & 50000 & 3072 & 50 & 1.5 & 1.9 & 2.9 & 1.0 & 1.2 & - & \textbf{9.4} \\
& & & 100 & 1.9 & 2.1 & 3.3 & 1.0 & 1.3 & - & \textbf{15.6} \\
& & & 200 & 2.3 & 2.1 & 3.4 & 1.0 & 1.2 & - & \textbf{26.2} \\
& & & 500 & 4.6 & 2.2 & 3.7 & 1.0 & 1.3 & - & \textbf{53.5} \\
& & & 1000 & 5.6 & 2.3 & 3.9 & 1.0 & 1.3 & - & \textbf{81.7} \\
cnnvoc & 15662 & 4096 & 50 & \textbf{15.5} & 1.5 & 2.1 & 1.0 & 1.3 & - & 8.1 \\
& & & 100 & \textbf{27.6} & 1.5 & 2.1 & 1.0 & 1.2 & - & 11.9 \\
& & & 200 & \textbf{41.0} & 1.5 & 2.2 & 1.0 & 1.3 & - & 19.5 \\
& & & 500 & \textbf{55.9} & 1.5 & 2.3 & 1.0 & 1.2 & - & 29.9 \\
& & & 1000 & 4.5 & 1.5 & 2.4 & 1.0 & 1.1 & - & \textbf{28.2} \\
covtype & 150000 & 54 & 50 & - & 4.2 & - & 1.0 & - & - & \textbf{23.9} \\
& & & 100 & - & 4.2 & - & 1.0 & - & - & \textbf{36.4} \\
& & & 200 & - & 4.4 & - & 1.0 & - & - & \textbf{61.2} \\
& & & 500 & - & 5.1 & - & 1.0 & - & - & \textbf{123.9} \\
& & & 1000 & - & 5.5 & - & 1.0 & - & - & \textbf{154.9} \\
mnist & 60000 & 784 & 50 & \textbf{14.1} & 2.3 & 3.5 & 1.0 & 1.4 & 1.4 & 10.0 \\
& & & 100 & \textbf{16.8} & 2.2 & 3.4 & 1.0 & 1.2 & 1.0 & 12.4 \\
& & & 200 & \textbf{30.0} & 2.4 & 3.9 & 1.0 & 1.2 & 0.3 & 20.6 \\
& & & 500 & \textbf{67.5} & 2.6 & 4.4 & 1.0 & 1.2 & - & 35.1 \\
& & & 1000 & \textbf{87.8} & 2.6 & 4.5 & 1.0 & 1.0 & - & 42.4 \\
mnist50 & 60000 & 50 & 50 & \textbf{12.9} & 2.3 & 3.4 & 1.0 & 1.2 & 1.0 & 6.6 \\
& & & 100 & \textbf{18.3} & 2.6 & 4.2 & 1.0 & 1.3 & 1.1 & 11.6 \\
& & & 200 & \textbf{23.2} & 2.8 & 4.6 & 1.0 & 1.2 & 0.8 & 19.2 \\
& & & 500 & 2.8 & 2.9 & 5.0 & 1.0 & 1.1 & 0.2 & \textbf{31.0} \\
& & & 1000 & 4.1 & 2.9 & 5.0 & 1.0 & 1.0 & - & \textbf{36.2} \\
tinygist10k & 10000 & 384 & 50 & \textbf{20.3} & 1.7 & 2.7 & 1.0 & 1.4 & - & 10.3 \\
& & & 100 & \textbf{24.2} & 1.7 & 2.6 & 1.0 & 1.3 & - & 15.1 \\
& & & 200 & \textbf{33.3} & 1.8 & 2.8 & 1.0 & 1.3 & - & 21.8 \\
& & & 500 & 2.9 & 1.7 & 2.8 & 1.0 & 1.2 & - & \textbf{30.6} \\
& & & 1000 & 2.3 & 1.7 & 2.7 & 1.0 & 0.7 & - & \textbf{23.7} \\
usps & 7291 & 256 & 50 & 9.5 & 2.7 & 3.7 & 1.0 & 0.6 & 0.2 & \textbf{9.7} \\
& & & 100 & 10.5 & 3.3 & 5.2 & 1.0 & 0.8 & - & \textbf{15.1} \\
& & & 200 & \textbf{30.1} & 3.3 & 5.3 & 1.0 & 0.5 & - & 21.5 \\
& & & 500 & \textbf{26.7} & 2.7 & - & 1.0 & - & - & 23.5 \\
& & & 1000 & \textbf{25.2} & 2.2 & - & 1.0 & - & - & 11.6 \\
yale & 2414 & 32256 & 50 & 2.2 & 3.0 & 4.9 & 1.0 & 0.9 & - & \textbf{14.2} \\
& & & 100 & 1.7 & 2.3 & 3.7 & 1.0 & 0.4 & - & \textbf{12.9} \\
& & & 200 & \textbf{28.3} & 2.3 & - & 1.0 & - & - & 14.9 \\
& & & 500 & \textbf{29.6} & 2.0 & - & 1.0 & - & - & - \\
& & & 1000 & - & \textbf{1.6} & - & 1.0 & - & - & - \\

\hline
\end{tabular}
\end{center}
\caption{Algorithmic speedup in reaching an energy within $2\%$ from the final Lloyd++ energy. (-) marks failure in reaching the target of $2\%$ relative error.
For each method, the parameter(s) that gave the highest speedup at 2\% error is used. }
\label{tab:k2means_speedup20_exp}
\end{table*}
}

\end{document}